# Hateful Person or Hateful Model? Investigating the Role of Personas in Hate Speech Detection by Large Language Models

**Shuzhou Yuan**[*1], **Ercong Nie**[*2,3], **Mario Tawfelis**[*1],
**Helmut Schmid**[2], **Hinrich Schütze**[2,3] and **Michael Färber**[1]
[1]ScaDS.AI and TU Dresden  [2]LMU Munich
[3]Munich Center for Machine Learning (MCML)
shuzhou.yuan@tu-dresden.de, nie@cis.lmu.de

## Abstract

**Content Warning:** *This paper contains examples of hate speech, which may be disturbing or offensive to some readers.*

Hate speech detection is a socially sensitive and inherently subjective task, with judgments often varying based on personal traits. While prior work has examined how socio-demographic factors influence annotation, the impact of personality traits on Large Language Models (LLMs) remains largely unexplored. In this paper, we present the first comprehensive study on the role of persona prompts in hate speech classification, focusing on MBTI-based traits. A human annotation survey confirms that MBTI dimensions significantly affect labeling behavior. Extending this to LLMs, we prompt four open-source models with MBTI personas and evaluate their outputs across three hate speech datasets. Our analysis uncovers substantial persona-driven variation, including inconsistencies with ground truth, inter-persona disagreement, and logit-level biases. These findings highlight the need to carefully define persona prompts in LLM-based annotation workflows, with implications for fairness and alignment with human values.

## 1 Introduction

The proliferation of hate speech on online platforms poses a persistent threat to inclusive digital spaces, necessitating robust and scalable detection systems (Röttger et al., 2021; Fortuna et al., 2022). Traditionally, the effectiveness of automated hate speech detectors has hinged on the quality of human-labeled data (Kim et al., 2022; Yuan et al., 2022). However, hate speech annotation is inherently subjective: as shown in Figure 1, what one annotator deems hateful, another may consider benign, with disagreement often rooted in personal beliefs, backgrounds, and identities (Sap et al., 2022; Fleisig et al., 2023; Astorino et al., 2023). Recent research has highlighted that socio-demographic factors such as age, gender, and cultural background significantly influence how annotators perceive and label hate speech, leading to biases that can propagate into downstream models (Mostafazadeh Davani et al., 2022; Wang and Plank, 2023).

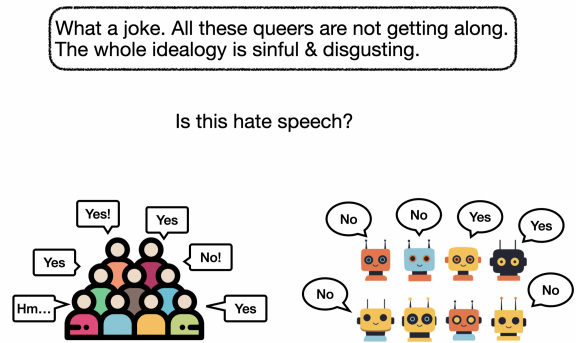

Figure 1: Our study demonstrates that different personas influence hate speech detection for both humans and large language models.

Large Language Models (LLMs) can be prompted to perform a wide range of classification, generation, and evaluation tasks (Zhang et al., 2023; Yang et al., 2024b; Wu et al., 2024). Recent work has demonstrated that LLMs, when provided with clear task instructions, are capable of matching or even surpassing the average accuracy of human annotators in a variety of labeling tasks, including hate speech detection, natural language inference, and sentiment analysis (Tan et al., 2024; Movva et al., 2024; Horych et al., 2025). As a result, *LLM-as-a-annotator* paradigms are being actively explored as alternatives or supplements to traditional human labeling pipelines (He et al., 2024). However, recent studies question the reliability of LLM annotations and indicate that LLMs can exhibit significant annotation variation (Borah and Mihalcea, 2024; Gligoric et al., 2025). To systematically study and control this variation, researchers have introduced persona-based prompting (Zhu et al., 2025), a technique that conditions LLMs to emulate specific annotator identities, backgrounds, or psychological profiles (Deshpande et al., 2023; Joshi et al., 2024). For example, LLMs can be prompted to annotate text as if they were a "young adult from Europe", a "conservative American", or a "first-generation immigrant", allowing researchers to probe how such identities influence model behavior (Joshi et al., 2024; Hu and Collier, 2024; Liu et al., 2024). This approach not only helps to audit for biases and blind spots within LLMs but also offers a pathway to generate more diverse and representative annotation data (Fröhling et al., 2024; Prpa et al., 2024).

---
*Equal contribution.

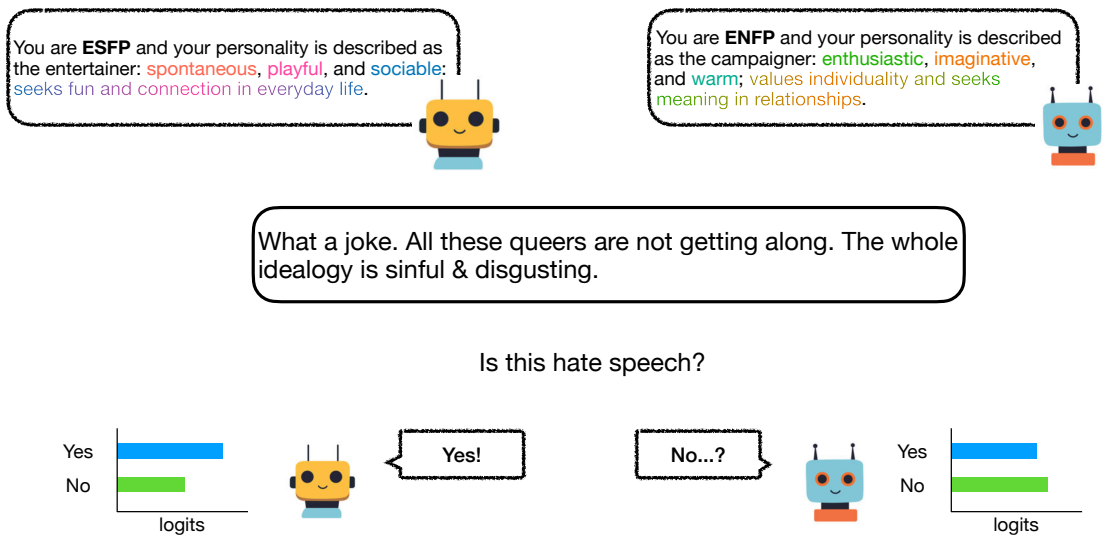

Figure 2: Assigning different MBTI-based personas to the LLM results in varied responses in hate speech detection. The predicted logits reveal clear differences in classification tendencies and confidence across personas. Notably, even a single personality dimension difference between ESFP and ENFP leads to divergent model outputs.

Despite its promise, most persona-based LLM research has focused on socio-demographic factors (e.g., age, gender, geographic background), with limited attention given to more fundamental and personally subjective traits such as personality (Orlikowski et al., 2023; Masud et al., 2024; Giorgi et al., 2024). In psychology and the social sciences, personality frameworks like the Myers–Briggs Type Indicator (MBTI) have long been used to explain individual differences in perception, judgment, and decision-making (Myers and Myers, 2010). The MBTI theory cat-egorizes individuals along four dichotomous dimensions (e.g., Introversion-Extraversion, Judging-Perceiving), which together are hypothesized to shape core aspects of a person's worldview and evaluative tendencies. In the NLP community, MBTI and similar psychological frameworks such as the Big Five Personality model (Goldberg, 2013) have been used both for personality detection from text (Plank and Hovy, 2015; Stajner and Yenikent, 2021; Li et al., 2025) and, more recently, to construct LLM personas for controlled generation and analysis (Jiang et al., 2024).

Yet, a systematic investigation of how personality-based personas influence subjective annotation tasks, such as hate speech detection, remains lacking. Most existing work treats personality as a downstream application or as a tool for behavioral simulation (Cheng et al., 2023; Lee and Ram, 2024; Choi and Li, 2024), rather than as a source of annotation variation in its own right. This raises important open questions: To what extent does personality drive annotator disagreement in subjective NLP tasks? Can LLMs, when prompted with different personality profiles, replicate or even amplify these effects? And what are the implications for the reliability, fairness, and interpretability of LLM-generated annotations?

In this work, we present the first comprehensive study investigating the influence of MBTI-based personas on hate speech detection by LLMs. We begin with a human annotation survey, in which participants report their MBTI personality types and annotate a curated set of hate speech examples. The results reveal that MBTI personality traits significantly affect individual judgments, particularly highlighting consistent differences between the *Feeling* and *Thinking* types. Building on this insight, we apply persona-based prompting to four open-source LLMs across three hate speech datasets. Our analysis shows substantial disagreement between model predictions and the original dataset labels. While each model displays distinct patterns of bias, we also observe notable variability in predictions across different personas within the same model. Further, as shown in Figure 2, a logit-level analysis reveals that even small changes in MBTI dimensions can lead to systematic shifts in model confidence and decision boundaries. When comparing human behavior with LLM outputs, we find that certain personality traits are more exaggerated in LLMs, indicating an amplification effect induced by persona prompts.

Our study leads to the following key insights:

- Human annotators with different MBTI personas produce significantly divergent hate speech annotations, and LLMs, when conditioned on persona prompts, also exhibit varied behavior.

- Persona conditioning not only affects final classification outcomes but also alters the model's internal confidence, revealing subtle but systematic influences on decision-making.

- In LLMs, analytical traits (T and J) increase confidence in hate speech predictions, contrasting with human data where Feeling types label more content as hateful.

- LLMs are more susceptible to persona framing than humans, amplifying certain behavioral traits and highlighting the risks of unintended bias in sensitive tasks.

## 2 Related Work

**Human and LLM Annotation Variation** Prior work has highlighted that human label variation in NLP tasks is not mere annotation noise, but often reflects meaningful signals such as linguistic ambiguity or subjective interpretation (Aroyo and Welty, 2013; Uma et al., 2021; Plank, 2022). Research shows that annotator disagreement is particularly pronounced in subjective tasks, motivating frameworks such as perspectivism and descriptive annotation to better capture inherent variation (Röttger et al., 2022; Cabitza et al., 2023). While many studies examine how sociodemographic factors influence annotation behavior (Fornaciari et al., 2021; Sap et al., 2022; Goyal et al., 2022), recent findings indicate that substantial individual variation remains unexplained by demographics alone (Orlikowski et al., 2023). Parallel lines of work have begun to benchmark the alignment between LLM and human annotations, focusing on group-level and pluralistic perspectives (Sorensen et al., 2024; Movva et al., 2024). However, little attention has been given to the role of psychological factors such as personality traits in shaping both human and LLM annotation variation.

**Hate Speech Detection** Hate speech detection has traditionally relied on two main strategies: leveraging supplementary user or annotator information and applying advanced language models fine-tuned on hate speech datasets (Nirmal et al., 2024). While utilizing user attributes or annotator traits can improve detection accuracy, such data are often difficult to obtain across platforms (Kim et al., 2022; Yin et al., 2023; del Valle-Cano et al., 2023; Waseem and Hovy, 2016). More commonly, language models like BERT and its variants are employed to generalize from large text corpora, with additional fine-tuning boosting their ability to recognize nuanced or implicit hate speech (Caselli et al., 2021; Mathew et al., 2021; Yuan et al., 2022). In this work, we focus on the subjective nature of hate speech detection by examining how LLM- and human-annotator variation, conditioned on personality traits, influences labeling patterns in this domain.

**Persona-based Prompting of LLMs** With the use of persona-based prompting, assigning specific roles or identities to guide model behavior, LLMs have been employed to generate human-like behavior in reasoning (Binz and Schulz, 2023; Ziems et al., 2024), role-playing (Wang et al., 2024a, 2025), social science experiments (Horton, 2023; Park et al., 2023; Wang et al., 2024b), and data annotation or evaluation (Ge et al., 2024; Dong et al., 2024). However, few studies have systematically examined how persona-based prompting, especially with psychologically grounded traits such as MBTI profiles, influences LLM annotation behavior and contributes to subjective variation in tasks like hate speech detection.

## 3 Human Survey: MBTI and Hate Speech

To investigate how MBTI personality types affect human judgment of hate speech, we conduct a survey and invite participants to label hate speech based on their MBTI personality.

According to the MBTI instrument, four dichotomous dimensions classify individuals as either extroverted (E) or introverted (I), sensing (S) or intuitive (N), thinking (T) or feeling (F), and judging (J) or perceiving (P) (Boyle, 1995). These four dichotomies result in sixteen possible combinations, which in turn lead to sixteen distinct personality types.[1]

### 3.1 Survey Design

We conduct an anonymous online survey to gain insight into how humans with different MBTI personalities perceive hate speech and offensive language. We select a subset of 20 samples from a commonly used hate speech dateset (Davidson et al., 2017). It contains 10 samples classified as hate speech and 10 samples classified as offensive language but not as hate speech in the original annotation.[2] Participants are prompted with one sample at a time, led by the question "Is this text hate speech?", and have the option to select either "Yes" or "No". After classifying all samples, participants are requested to provide information about their MBTI personality type as well as their demographic information.

### 3.2 Survey Results and Analysis

We distribute the survey through the university's mailing list, targeting students and staff from various academic disciplines. A total of 293 valid responses are collected, representing all 16 MBTI personality types. The survey results are compared against the ground truth labels from the original dataset, in which 50% of the samples are annotated as hate speech.

Figure 3 presents the average percentage of samples labeled as hate speech ("Yes") by each MBTI

---

[1] The description of the personas can be found in Appendix C.

[2] The selection of the samples is not random as we want to make sure the text is meaningful and not only contains slang or user tagger. The details of the survey can be found in Appendix B.

personality. All personality types show variation from the 50% hate speech rate found in the ground truth. ESFP labels the highest proportion of samples as hate speech (67.5%), followed by INFJ, ENFP, ISFJ, and INFP, all of whom label over 60% of samples as hate speech. These personalities all include the Feeling (F) trait, which may reflect a greater sensitivity to offensive content. On the other hand, ISTJ, ESFJ, and ESTP label fewer than 50% of the samples as hate speech, indicating a more lenient or conservative interpretation. Two of these types show the Thinking (T) trait, which may correspond to a more analytical or detached judgment process. *These findings suggest that individual personality traits, particularly the Feeling–Thinking dimension, systematically influence how people perceive and classify hate speech.*

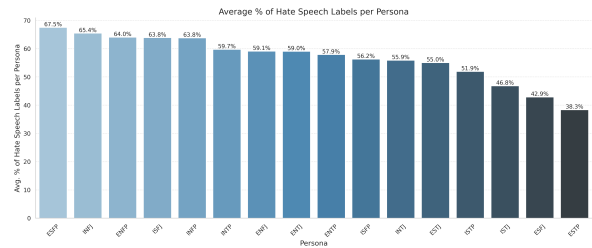

Figure 3: Average percentage of hate speech labels by 16 MBTI persona type in the human survey.

## 4 Task Formulation for Hate Speech Classification

To investigate if LLMs also inherit the subjectivity of the judgment to hate speech from human, we conduct experiments on LLMs by giving them different persona roles. As shown in Figure 2, the LLM is defined with different persona prompt and given a potential toxic text to answer if it belongs to hate speech, the model processes the input and generates the next token as the predicted label $\hat{y}$. We constrain the output vocabulary space by instructing the model to "only answer 'Yes' or 'No'" for binary classification tasks. The predicted label is determined by selecting the token with the highest probability:

$$\hat{y} = \arg\max_{y \in V} P_{\phi}(y) \tag{1}$$

where $P_{\phi}(y)$ represents the generative probability from the LLM, and $V$ is the vocabulary space for the labels, defined as $V = \{\text{Yes, No}\}$ in our tasks. We use the generated label word to analyze the inconsistency between the ground truth and the label generated by the LLM. Additionally, we use the logits of the generated label words 'Yes' and 'No' to analyze inconsistencies between different personas, as well as the tendency or bias toward specific dimensions.

The persona of LLMs is defined by providing the persona information in the system prompt:

> *System*: You are $[Persona]$ and your personality is described as $[Description]$.
>
> *User*: $[Text]$ Is this text hate speech? Only answer with one word, yes or no.

where $[Persona]$ indicate the 16 personalities in MBTI, $[Description]$ is the official description about the personality, and $[Text]$ is the sample from the hate speech datasets.[3]

## 5 Experiments

### 5.1 Dataset

| Dataset | Number of Samples |
|---|---|
| CREHate | 365 |
| HateXplain | 1142 |
| Davidson | 20620 |

Table 1: Statistics of the hate speech datasets used in this research.

An overview of the datasets used in this study is provided in Table 1. We use three widely adopted hate speech datasets:

**CREHate** (Lee et al., 2024) The **CR**oss-cultural **E**nglish **Hate** speech dataset contains 1,580 samples which have been sourced from social media platforms. Human annotators, classifying text as hate speech or otherwise, were selected from five English-speaking countries. In our study, we use samples in which all countries unanimously classified as hate speech.

**HateXplain** (Mathew et al., 2021) The HateXplain dataset, compromised of around 20k samples, classifies text in three groups: hateful, offensive, normal, or undecided. Using only samples that are classified as hate speech or offensive language, we take approximately 1.1k samples for this research.

**Davidson** (Davidson et al., 2017) is a dataset of approximately 29k samples sourced from Twitter. Human annotators were asked to label the samples into three categories: hate speech, offensive language (not hate speech), or neither hate speech nor offensive. In our project, we only consider samples that are either classified as hate speech or offensive language, which results in approximately 20k samples.

### 5.2 Models and Setup

We use four open source instruction-tuned LLMs with the size of 7-8B in our experiments: `Llama-3.1-8B` (Dubey et al., 2024), `Ministral-8B` (MistralAI, 2024), `Falcon3-Mamba-7B` (Zuo et al., 2024), and `Qwen2.5-7B` (Yang et al., 2024a).[4] We set the generation temperature to 0 to keep a deterministic output with no randomness.

## 6 Results and Analysis

### 6.1 Inconsistency between LLM and Human Labels

The results of hate speech classification with different MBTI-based personas for four LLMs are presented in Figure 4. We report the inconsistency percentage, defined as the proportion of instances where the LLM-generated label diverges from the ground truth label. The results reveal noticeable variation across datasets, models, and persona types.

Across all three datasets, CREHate, HateXplain[5], and Davidson, we observe that the choice of LLM significantly influences inconsistency rates. For example, in the CREHate dataset (Figure 4a), `Ministral-8B-Instruct` consistently shows the highest inconsistency across all personas, often reaching 70% or higher. In contrast, `Qwen2.5-7B-Instruct` and `Llama3.1-8B-Instruct` exhibit substantially lower inconsistency, particularly for Thinking (T) and Judging (J) persona types such as ENTJ and ISTJ. `Mamba-7B-Instruct` maintains moderate inconsistency levels, typically between 30% and 50%.

The Davidson dataset (Figure 4b) reveals the highest overall inconsistency rates, with almost all models frequently exceeding 70% inconsistency across most personas. `Llama3.1-8B-Instruct` demonstrates particularly high divergence from the ground truth, while

---

[3]The personas and the corresponding description can be found in Appendix C.

[4]The details of the LLMs can be found in Appendix A.

[5]The result of HateXplain is presented in Appendix D Figure 10, as it presents more uniform inconsistency levels across LLMs.

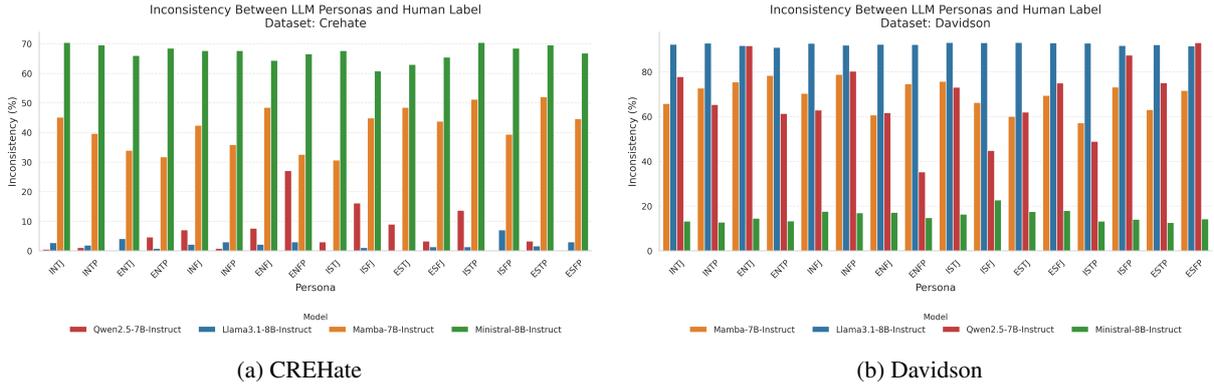

(a) CREHate                                   (b) Davidson

Figure 4: Inconsistency between LLM predictions and ground truth labels across 16 personas on three hate speech datasets for four LLMs.

`Ministral-8B-Instruct` achieves relatively better alignment. `Qwen2.5-7B-Instruct` exhibits not only high inconsistency with the ground truth but also substantial disagreement across different personas.

*Overall, these results demonstrate that both the choice of LLM and the assigned MBTI persona significantly influence hate speech classification, with the degree of impact varying across datasets.*

## 6.2 Intra-Model Disagreement Across Personas

As the LLMs exhibit inconsistencies with the ground truth labels, we further investigate the disagreements among different MBTI personas conditioned within the same LLM. Figure 5 presents the pairwise disagreement matrix across all 16 MBTI personas for `Qwen2.5-7B-Instruct` on the Davidson dataset, as Davidson shows the highest overall inconsistency between model predictions and human labels (see § 6.1).[6]

Each cell in the matrix quantifies the disagreement rate, ranging from 0 to 1, where higher values indicate more frequent divergence in predicted labels on the same inputs. ENFP emerges as the most divergent persona, with disagreement values around 0.6 with multiple others, including ISTJ (0.57), ESFP (0.63), and ENTJ (0.61). In contrast, logical and structured personas such as ISTJ, INTJ, and ESTJ show lower pairwise disagreement among themselves, such as INTJ–ISTJ (0.08) and ESTJ–ISTJ (0.13), indicating more aligned classification behavior.

Feeling and Perceiving types such as INFP, ESFP, and ENFP exhibit higher disagreement lev-

⁶Disagreement matrices for other datasets and LLMs can be found in Appendix E.

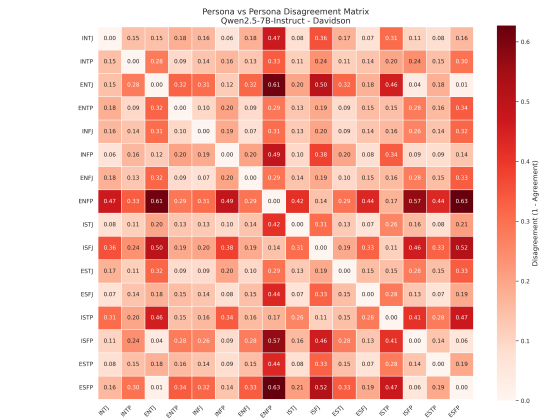

Figure 5: The Persona vs Persona Disagreement Matrix for `Qwen2.5-7B-Instruct` on Davidson, higher value denotes higher disagreement. Diagonal values are zero by design, as they represent self-agreement.

els with others, suggesting that more emotionally driven personas introduce greater variability in model predictions. The highest disagreement is observed between ENFP and ESFP (0.63), despite both being extraverted and feeling-oriented, suggesting that even small trait differences such as Intuition vs. Sensing can significantly alter the model's labeling decisions.

*Overall, these results reveal that intra-model disagreement is strongly influenced by MBTI traits, with emotionally driven (F) and less structured (P) personas introducing the highest variability, while analytical (T) and judging (J) types yield more consistent predictions across personas.*

## 6.3 Tendency Analysis from Logits

The intra-model disagreement shows that persona plays a significant role in hate speech judgment. We further analyze the underlying decision tendencies by examining the logits produced by the LLMs for the 'yes' and 'no' responses. Specifically, we investigate whether LLMs exhibit a systematic bias toward one label over the other when conditioned on different personas.

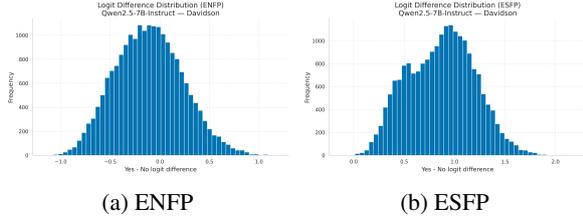

(a) ENFP       (b) ESFP

Figure 6: Logit Difference Distribution of `Qwen2.5-7B-Instruct` for ENFP vs ESFP on Davidson. The results for other personas, datasets and LLMs can be found in Appendix I.

Given that ENFP and ESFP demonstrate high disagreement on `Qwen2.5-7B-Instruct` for the Davidson dataset, as discussed in § 6.2, we analyze the logit difference distributions for these two personas. Figure 6 presents the distribution of the difference between the logit for the 'yes' token and the logit for the 'no' token, computed as $\text{logit}_{yes} - \text{logit}_{no}$.

A clear contrast is observed between the two distributions. For ESFP (Figure 6b), the logit differences are strictly greater than zero, indicating that the model consistently assigns higher confidence to the 'yes' class. This suggests a strong inclination toward predicting that the input is hate speech under the ESFP persona. In contrast, for ENFP (Figure 6a), the logit differences are more evenly distributed around zero, with a slight skew toward the negative side, indicating that the 'no' token tends to receive slightly higher scores than 'yes'. This pattern suggests a general bias toward predicting non-hate speech when the model is conditioned on the ENFP persona.

*These results suggest that persona traits influence not only the final classification outcomes but also the model's internal decision confidence.* While the ESFP persona, characterized by expressiveness and sensitivity to external stimuli, leads to a strong bias toward labeling inputs as hate speech, the ENFP persona, often associated with empathy and open-mindedness, shows a more balanced, slightly non-hate-leaning tendency.

### 6.4 MBTI Dichotomies Logit Distributions

As ENFP and ESFP show clear logit-level tendencies despite differing in only one MBTI dimension, we further examine how each individual MBTI dichotomy, Extraversion (E) vs. Introversion (I), Intuition (N) vs. Sensing (S), Thinking (T) vs. Feeling (F), and Perceiving (P) vs. Judging (J), affects model behavior in hate speech classification. We focus on the "yes" logit values, which represent the model's internal confidence in predicting hate speech, and analyze their distribution across each dichotomy. The results for all four LLMs on the Davidson dataset are shown in Figure 7.[7]

In the Extraversion vs. Introversion comparison (Figure 7a), we observe that extraverted personas tend to produce slightly higher "yes" logits than introverted ones for `Qwen` and `Ministral`, while introverted personas tend to produce slightly higher "yes" logits for `Llama`. However, the difference is not very evident for this group. In the Intuition vs. Sensing comparison (Figure 7b), sensing types show a modest shift toward higher "yes" logits. This pattern is relatively consistent across models and may suggest that sensing personas are more sensitive to concrete or literal hate indicators in the text. For Thinking vs. Feeling (Figure 7c), thinking personas consistently yield higher "yes" logits than feeling ones across all models. The effect is especially pronounced in `Qwen` and `Llama`, which suggests that personas emphasizing analytical reasoning are more likely to assign stronger confidence scores when labeling inputs as hate speech. In the Perceiving vs. Judging dimension (Figure 7d), judging types generally produce higher "yes" logits than perceiving types. This trend, consistent across `Mamba`, `Qwen`, and `Llama`, could reflect a stronger preference among judging personas for making firm decisions regarding moral or ethical boundaries, including the detection of hate speech.

*Overall, these results suggest that even high-level personality traits can shape the internal decision-making dynamics of LLMs under persona conditioning. In particular, Thinking and Judging personas tend to produce stronger "yes" logits, reflecting a more decisive and analytical classification style.*

### 6.5 Comparison of Human and LLM Hate Speech Classification Patterns

We conduct PCA on the decision patterns of both human participants in §3 and a LLM (`Qwen2.5-7B-Instruct`) prompted with different persona descriptions. Figure 8 presents the PCA

---

[7]The results for other datasets can be found in Appendix H.

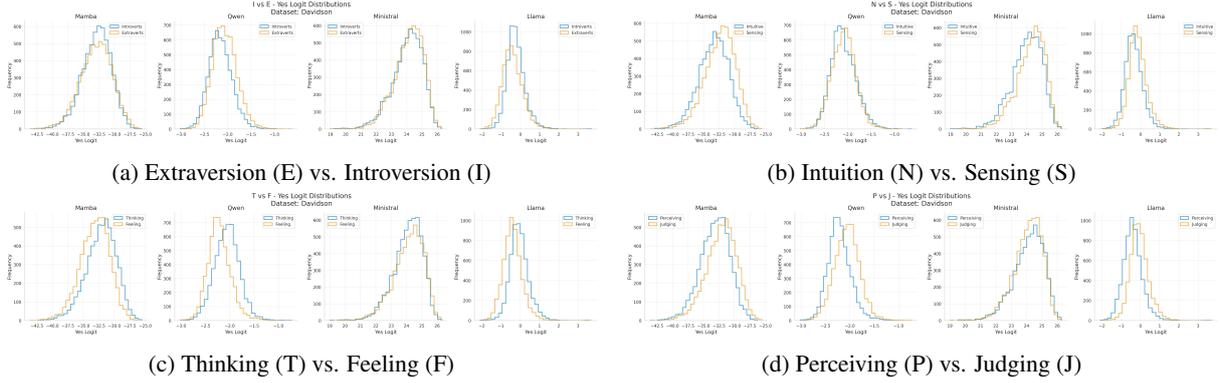

(a) Extraversion (E) vs. Introversion (I)

(b) Intuition (N) vs. Sensing (S)

(c) Thinking (T) vs. Feeling (F)

(d) Perceiving (P) vs. Judging (J)

Figure 7: "Yes" logit distributions for each MBTI dichotomy on the Davidson dataset, comparing behavior across four LLMs. Each subplot illustrates how different personality dimensions influence the model's confidence in classifying a sample as hate speech.

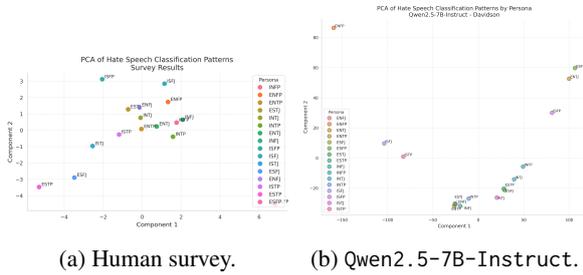

(a) Human survey.

(b) Qwen2.5-7B-Instruct.

Figure 8: PCA of hate speech classification for human survey and `Qwen`.

plots for human survey results and LLM predictions.[8]

The human PCA reveals a relatively compact and overlapping distribution of MBTI types. While some variation is observable, such as ESFP and ESTP appearing more distant from the central cluster, most personas form a coherent grouping. This suggests that although individual personality types influence judgment, human annotators share a broadly consistent understanding of what constitutes hate speech.

In contrast, the PCA of LLM predictions shows a starkly different pattern. The MBTI personas result in widely dispersed decision behaviors, with some types (e.g., ENFP, ESFP, ISFP) forming clear outliers far from the main cluster. This indicates that persona prompting has a significantly stronger impact on the LLM's decision boundary compared to the relatively subtle effects observed in human behavior.

One possible explanation for this divergence is that LLMs, when instructed to adopt a specific persona, may overfit to stereotypical traits described in the prompt. For example, a persona described as "caring, outgoing, and cooperative" might bias the model toward more lenient interpretations of controversial language. This sensitivity suggests that LLMs amplify personality cues more than humans naturally do, potentially leading to inconsistent or biased moderation outcomes.

*Overall, the comparison reveals that while human judgments are influenced but constrained by personalities, LLMs are far more malleable to persona prompt.*

# 7 Conclusion

In this work, we study the impact of MBTI-based personas on hate speech classification by combining a human survey with LLM analysis. Our survey reveals that personality traits influence labeling behavior, with significant variation across MBTI types. Extending this to LLMs, we find that persona prompts lead to substantial prediction shifts, including inconsistencies with ground truth and disagreements across personas. At the logit level, we observe systematic biases aligned with persona traits. Additionally, examining the influence of individual MBTI dichotomies reveals that Thinking and Judging traits are associated with stronger logit confidence in hate speech detection.

These findings demonstrate that persona conditioning in LLMs not only affects final predictions but also shapes the underlying decision-making process. As LLMs are increasingly used in socially sensitive applications, future work could explore how to better understand and control the influence of personas to ensure fairness, consistency, and reliability in subjective tasks.

---

[8]The results for other datasets and LLMs can be found in Appendices F and G.

## Limitations

While our study offers novel insights into persona-based prompting in LLMs, several limitations should be acknowledged. First, we focus exclusively on the MBTI framework as the basis for personality representation. Although MBTI is widely recognized and structured, future work could explore alternative psychological theories, such as the Big Five, to provide different perspectives. Second, our analysis is conducted on a limited set of benchmark hate speech datasets, which may not fully capture the richness and diversity of hate speech encountered in real-world settings. Third, we evaluate only four open-source LLMs, each with approximately 7 to 8 billion parameters. Future work should investigate whether similar effects occur in larger models. Fourth, our human survey sample consists primarily of students and researchers, which may limit the generalizability of our findings to other demographic groups. Finally, our investigation is restricted to English-language data. Cross-lingual studies are needed to examine how persona effects and perceptions of hate speech vary across different cultural and linguistic contexts.

## Ethical Considerations

This study investigates the influence of MBTI-based personas on LLM behavior in the context of hate speech detection. Given the sensitive nature of both hate speech content and personality profiling, several ethical considerations must be addressed.

**Hate Speech Content.** The use of hate speech examples, even for research purposes, carries the risk of exposing readers and researchers to harmful or offensive material. We use such content solely to support the scientific objectives of this work and explicitly disclaim any intent to promote or reproduce hateful language. A content warning is provided at the beginning of the paper to mitigate potential harm.

**Use of MBTI Personas.** While the MBTI framework has limited psychometric validity, we adopt it as a structured and interpretable method for simulating persona variation in LLMs. The MBTI profiles are used as fictional constructs to explore behavioral diversity in model outputs and are not intended to diagnose, stereotype, or represent real individuals. Our use of MBTI is exploratory in nature and aims to highlight potential variation, not prescribe personality-based approaches.

**Bias and Stereotyping.** There is a potential ethical risk in associating specific personas with differing behaviors in hate speech detection, which could inadvertently reinforce stereotypes. To mitigate this, we conduct our analysis with care and emphasize that the observed patterns reflect model behavior under specific prompt conditions. No personality type is pathologized or presented as superior or inferior. Our findings should be interpreted as insights into LLM response variability, not as generalizations about personality traits.

## A  Models

We use four open source instruction-tuned LLMs with the size of 7-8B in our experiments. Further information about each model can be found on the respective HuggingFace pages below:

- Llama3.1: https://huggingface.co/meta-llama/Llama-3.1-8B-Instruct

- Ministral: https://huggingface.co/mistralai/Ministral-8B-Instruct-2410

- Qwen2.5: https://huggingface.co/Qwen/Qwen2.5-7B-Instruct

- Mamba: https://huggingface.co/tiiuae/falcon-mamba-7b-instruct

## B  Survey

### B.1  Setup

Participants were first provided with a definition for hate speech and offensive language along with an example for each. Below is the text included in the survey's introduction:

**Hate speech** is defined as language used to express hatred towards a targeted group/individual based on protected attributes such as race, religion, sexual orientation, etc.

Example: "All [religious group]s are terrorists. Deport them all!"

**Offensive language** contains offensive terms but is not targeting any group in particular.

Example: "Stop acting like a bitch and grow up already."

The survey contained a total of 20 samples with 10 samples classified as hate speech and 10 samples classified as offensive language but not hate speech. Below are the samples included in the survey, starting with the hate speech samples and followed by the offensive language samples:

- #California is full of white trash who moved from #Oklahoma

- #faggots are MERELY a #scientific indicator that SOMETHING IS WRONG; Do not celebrate as "unique indiviuals" for they are #cancer

- @[USERNAME]: No need to thank me, killing niggers is it's own reward. Dis my pal from our graduation when we was 26

- #firefighter is a job for white trash

- @[USERNMAE] @[USERNAME] So you support "illegals" coming here and spreading disease among Americans Dream that bitch!

- @[USERNAME] what's this chinks email? I'm moving to China and slicing his throat

- #SomethingIGetAlot Are you... asian? black? Hawaiian? gay? retarded? drunk?

- #JesusChrist was STRAIGHT That's why the #faggots killed him. #PERIOD #SonOfGod

- RT @[USERNAME]: Well maybe you're a whore you stupid ugly homeless bitch

- They thought I wouldn't make it pussy nigga we made it

- no youre a fucking retard thats gonna get fucking date raped at a frat tonight. have fun with those repressed memories

- Man bitches love Beyonce....no punch line

- Then ya'll idolize Kanye when he wifed up a hoe. Lol

- RT @[USERNAME]: When someone's phone goes off in the cinema I want to jump across the seats and fly kick dat bitch.

- RT @[USERNAME]: I convinced hitler was a faggit, no bitches, no kids, all those leather uniforms, wanted to be alone with his closest men lock

- "@[USERNAME]: No text back? I don't give a fuck, I'll text another bitch" damn someone is hurt lol

- @[USERNAME]'s a faggot. I'm glad your chipped him.

- Bitches cheat on they bitch... But be SOOOOOO HURT when they bitch cheat back.

- You a dike cause yo man a pussy.

- "@RTNBA: Drakes new shoes that will be released by Nike/Jordan.... Yes, there's glitter on the shoes [URL]" ....dudes a fag

### B.2 Demographics

The table below summarizes the distribution of the ethnicity groups to which participants of our survey belong to. This information was not mandatory as part of our survey and was voluntarily provided by the participants.

| Ethnic Group | Count |
|---|---|
| White | 216 |
| Asian | 44 |
| Other | 11 |
| Middle Eastern or North African (MENA) | 10 |
| Hispanic or Latino/a/x | 8 |
| Multiracial / Two or more races | 4 |

Table 2: Distribution of participants by ethnic group.

## C Personas

Below are the sixteen personas along with their most common traits. *16personalities.com* use gamified names for each persona, such as The Mediator, The Debater, etc., which have also been included below for completeness. It is also worth mentioning that the models were provided with these names along with the traits.

- INFP – The Mediator: Creative, introspective, and empathetic; guided by inner values and harmony.

- ENFP – The Campaigner: Enthusiastic, imaginative, and warm; values individuality and seeks meaning in relationships.

- ENTP – The Debater: Quick-witted, curious, and argumentative; enjoys intellectual challenges and exploring new ideas.

- ESTJ – The Executive: Organized, decisive, and pragmatic; excels at managing people and systems.

- INTJ – The Architect: Strategic, independent, and analytical; has a vision for the future and works to achieve it.

- INTP – The Logician: Innovative, curious, and analytical; loves exploring abstract ideas and theories.

- ENTJ – The Commander: Bold, strategic, and assertive; natural leader who thrives on achievement and efficiency.

- INFJ – The Advocate: Idealistic, insightful, and reserved; driven by values and a desire to help others.

- ISFP – The Adventurer: Gentle, artistic, and adaptable; values personal freedom and expression.

- ISFJ – The Defender: Loyal, empathetic, and practical; quietly supportive and protective of others.

- ISTJ – The Logistician: Responsible, serious, and detail-oriented; values tradition and duty.

- ESFJ – The Consul: Caring, outgoing, and cooperative; values harmony and tradition in social settings.

- **ENFJ – The Protagonist:** Charismatic, altruistic, and inspiring; driven to lead and help others grow.

- **ISTP – The Virtuoso:** Practical, observant, and spontaneous; enjoys hands-on problem-solving.

- **ESTP – The Entrepreneur:** Energetic, action-oriented, and outgoing; enjoys living in the moment and taking risks.

- **ESFP – The Entertainer:** Spontaneous, playful, and sociable; seeks fun and connection in everyday life.

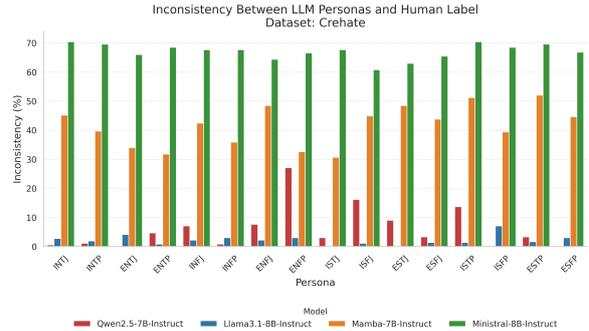

Figure 11: Percentage inconsistency between LLMs and the ground truth on the CReHate dataset

## D  Classification Inconsistency Between LLMs and GT

The figures below depict the percentage inconsistency between LLMs and ground truth for each persona.

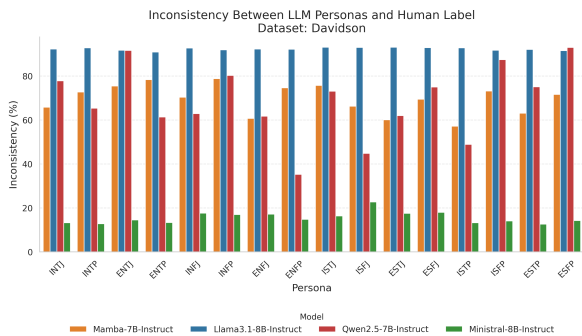

Figure 9: Percentage inconsistency between LLMs and the ground truth on the Davidson dataset

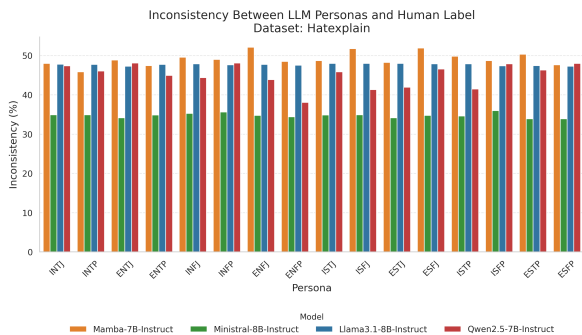

Figure 10: Percentage inconsistency between LLMs and the ground truth on the HateXplain dataset

## E  Disagreement Among Personas

The figures below depict the disagreement among personas.

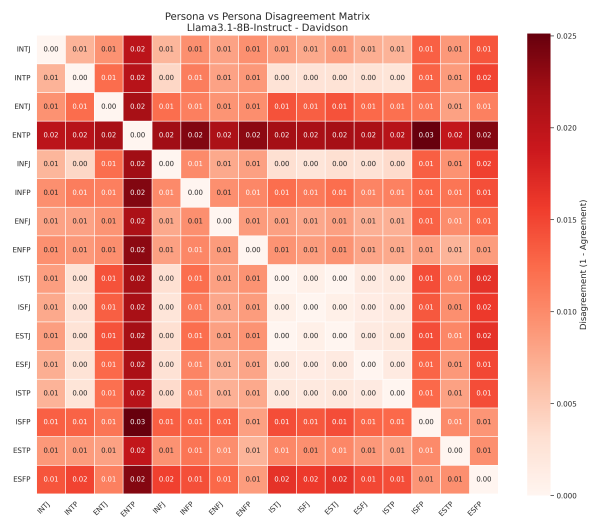

Figure 12: Disagreement among personas for `Llama3.1` on the Davidson dataset

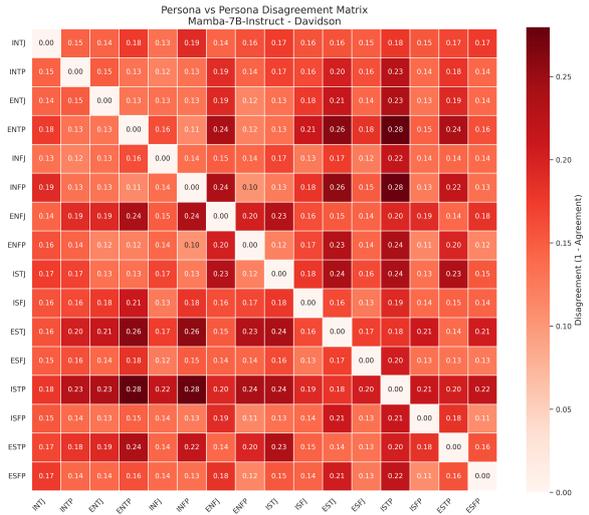

Figure 13: Disagreement among personas for `Mamba` on the Davidson dataset

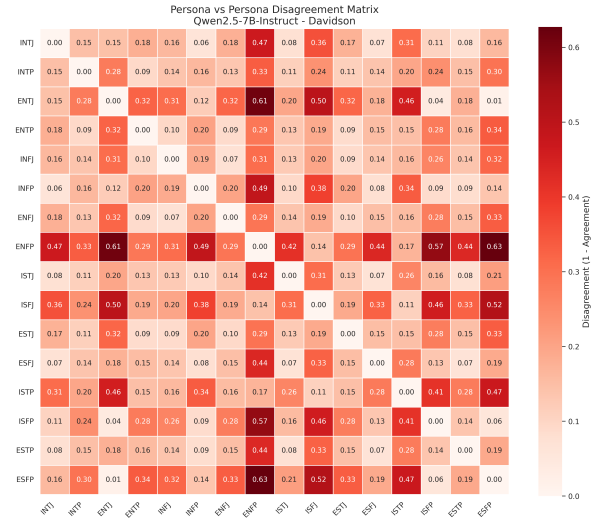

Figure 15: Disagreement among personas for `Qwen2.5` on the Davidson dataset

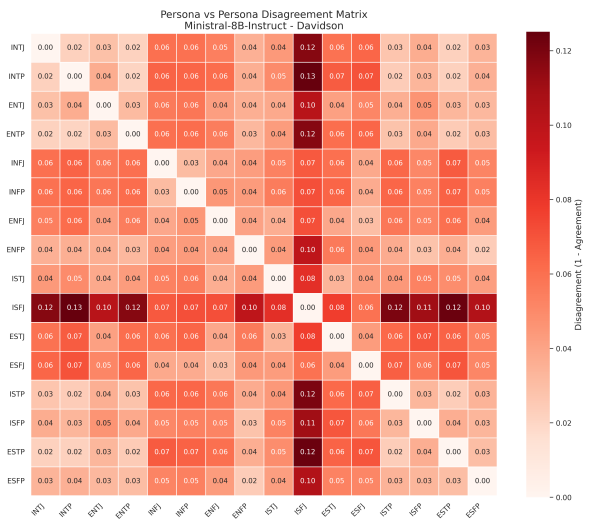

Figure 14: Disagreement among personas for `Ministral` on the Davidson dataset

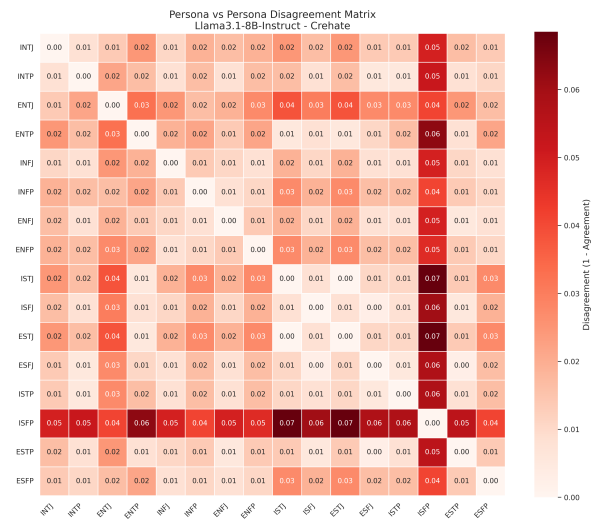

Figure 16: Disagreement among personas for `Llama3.1` on the CREHate dataset

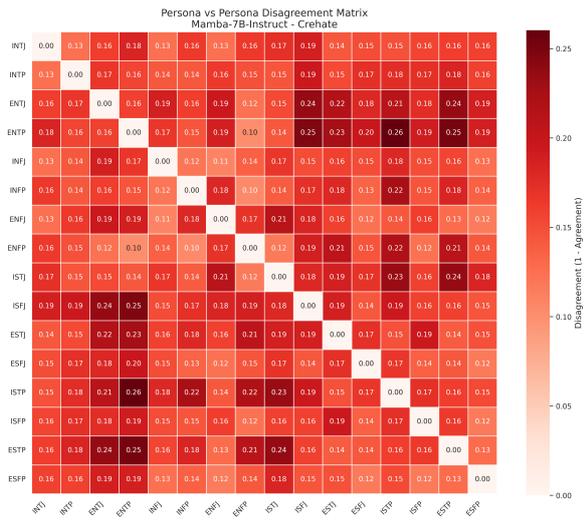

Figure 17: Disagreement among personas for `Mamba` on the CREHate dataset

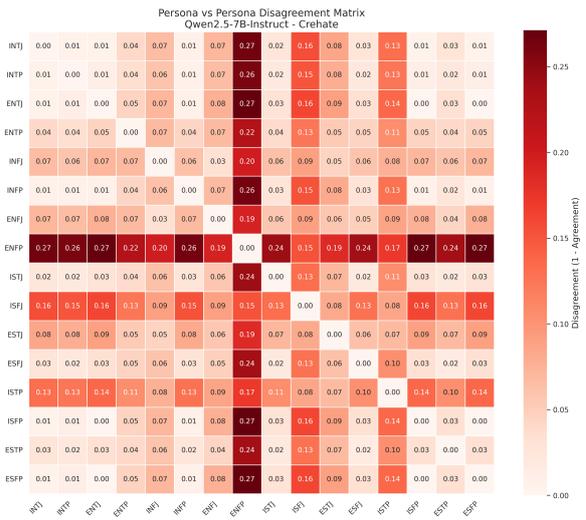

Figure 19: Disagreement among personas for `Qwen2.5` on the CREHate dataset

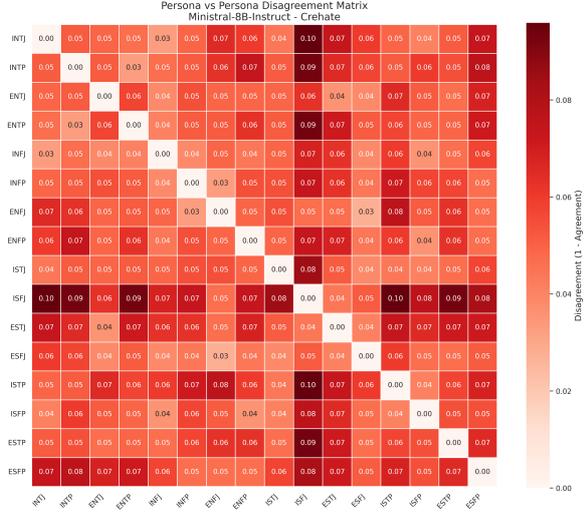

Figure 18: Disagreement among personas for `Ministral` on the CREHate dataset

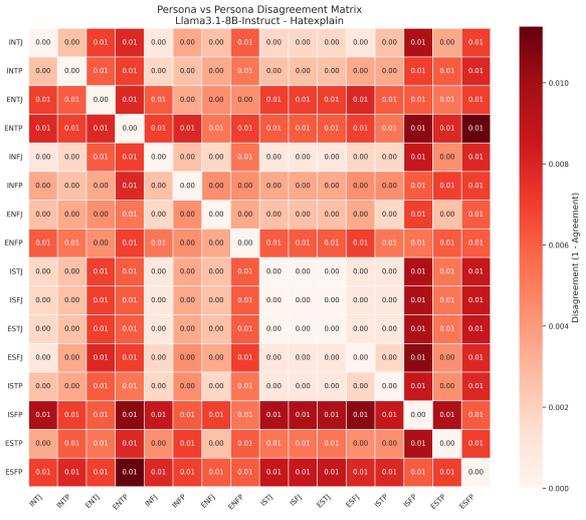

Figure 20: Disagreement among personas for `Llama3.1` on the HateXplain dataset

Figure 21: Disagreement among personas for `Mamba` on the HateXplain dataset

Figure 23: Disagreement among personas for `Qwen2.5` on the HateXplain dataset

# F   PCA

Figure 22: Disagreement among personas for `Ministral` on the HateXplain dataset

Figure 24: PCA of hate speech classification per persona for `Mamba` on the Davidson dataset

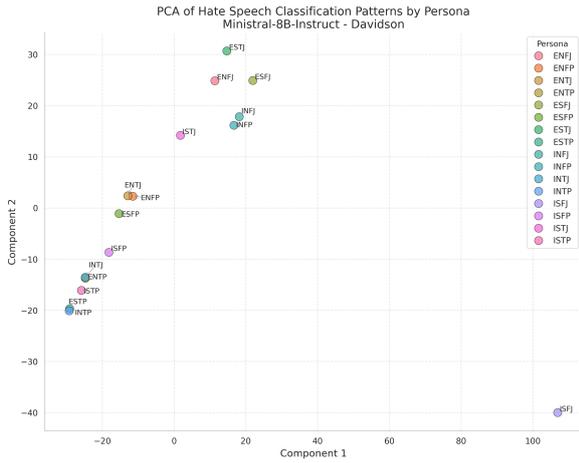

Figure 25: PCA of hate speech classification per persona for `Minisstral` on the Davidson dataset

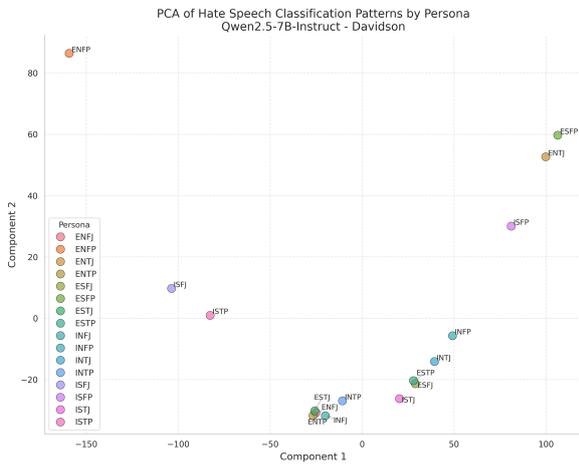

Figure 26: PCA of hate speech classification per persona for `Qwen2.5` on the Davidson dataset

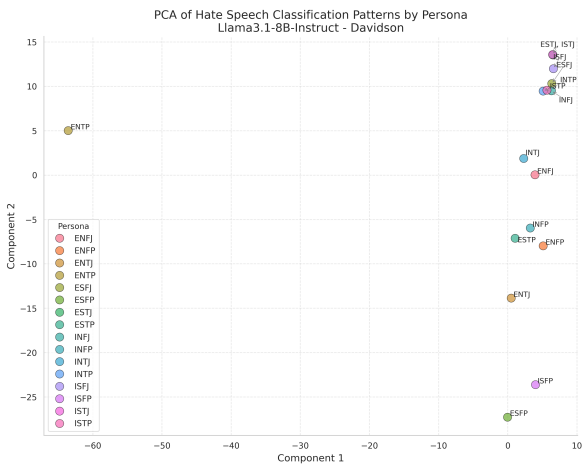

Figure 27: PCA of hate speech classification per persona for `Llama` on the Davidson dataset

## G Hierarchical Clustering

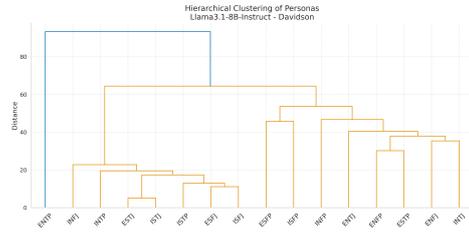

Figure 28: Hierarchical clustering of persona representations for `LLaMA3.1-8B-Instruct` on the Davidson dataset.

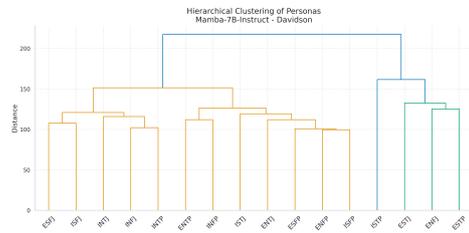

Figure 29: Hierarchical clustering of persona representations for `Mamba-7B-Instruct` on the Davidson dataset.

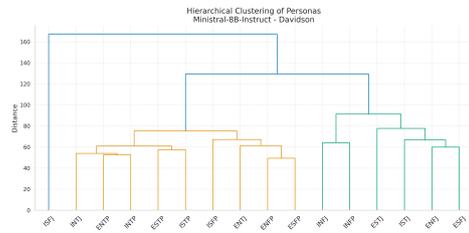

Figure 30: Hierarchical clustering of persona representations for `Ministral-8B-Instruct` on the Davidson dataset.

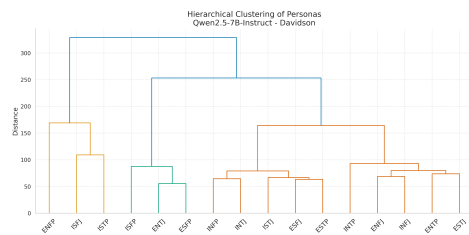

Figure 31: Hierarchical clustering of persona representations for `Qwen2.5-7B-Instruct` on the Davidson dataset.

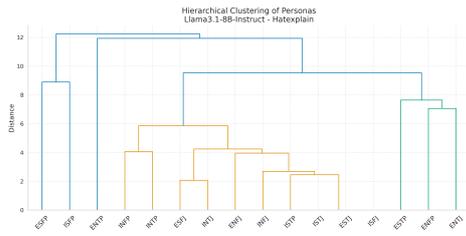

Figure 32: Hierarchical clustering of persona representations for `LLaMA3.1-8B-Instruct` on the HateXplain dataset.

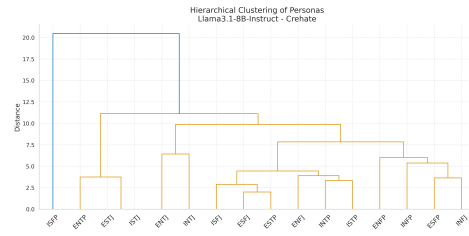

Figure 36: Hierarchical clustering of persona representations for `LLaMA3.1-8B-Instruct` on the CREHate dataset.

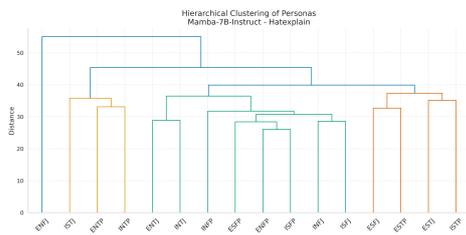

Figure 33: Hierarchical clustering of persona representations for `Mamba-7B-Instruct` on the HateXplain dataset.

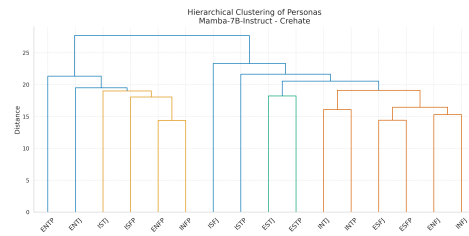

Figure 37: Hierarchical clustering of persona representations for `Mamba-7B-Instruct` on the CREHate dataset.

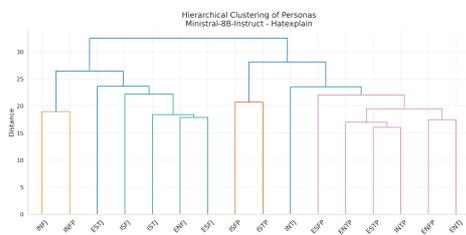

Figure 34: Hierarchical clustering of persona representations for `Ministral-8B-Instruct` on the HateXplain dataset.

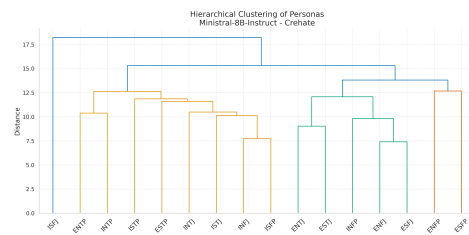

Figure 38: Hierarchical clustering of persona representations for `Ministral-8B-Instruct` on the CREHate dataset.

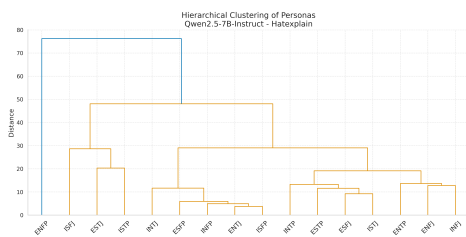

Figure 35: Hierarchical clustering of persona representations for `Qwen2.5-7B-Instruct` on the HateXplain dataset.

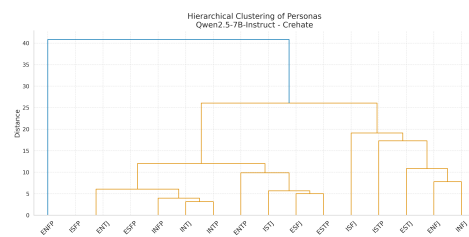

Figure 39: Hierarchical clustering of persona representations for `Qwen2.5-7B-Instruct` on the CREHate dataset.

## H 'Yes' Logit Distribution Per Dichotomy

In the context of this study, a dichotomy is a pair of persona groups with three fixed traits (e.g., Introversion, Thinking, Sensing, etc.) and one varying trait. It is defined, in our code, as such:

```
mbti_dichotomies = {
```

```
'I vs E': {
    'Introverts': ['INFP', 'ISFP', 'INTP
        ', 'INFJ'],
    'Extraverts': ['ENFP', 'ENTP', 'ESTJ
        ', 'ESFP']},
'N vs S': {
    'Intuitive': ['INFP', 'ENFP', 'INTP
        ', 'ENTP'],
    'Sensing': ['ISFP', 'ESTJ', 'ISFJ',
        'ISTJ']},
'T vs F': {
    'Thinking': ['INTJ', 'ENTJ', 'ISTJ',
        'ESTJ'],
    'Feeling': ['INFP', 'ENFP', 'ISFP',
        'ESFP']},
'P vs J': {
    'Perceiving': ['INFP', 'ENFP', 'ISFP
        ', 'ESFP'],
    'Judging': ['INTJ', 'ENTJ', 'ISTJ',
        'ESTJ']}
}
```

Below are the figures visualizing the 'Yes' logit distribution per dichotomy for each model on each dataset:

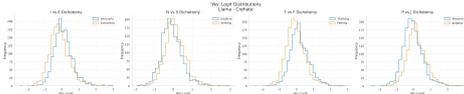

Figure 40: Distribution of yes logits for `LLaMA3.1-8B-Instruct` on the CREHate dataset.

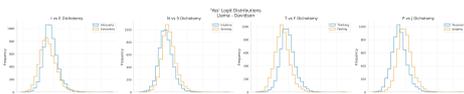

Figure 41: Distribution of yes logits for `LLaMA3.1-8B-Instruct` on the Davidson dataset.

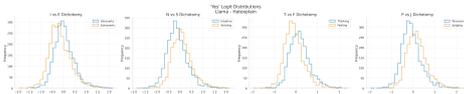

Figure 42: Distribution of yes logits for `LLaMA3.1-8B-Instruct` on the HateXplain dataset.

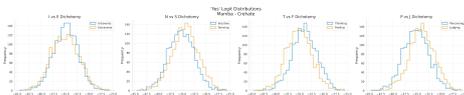

Figure 43: Distribution of yes logits for `Mamba-7B-Instruct` on the CREHate dataset.

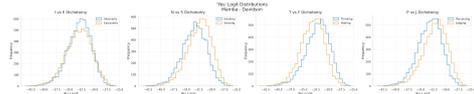

Figure 44: Distribution of yes logits for `Mamba-7B-Instruct` on the Davidson dataset.

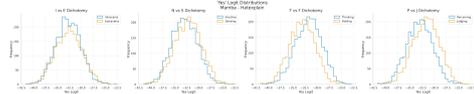

Figure 45: Distribution of yes logits for `Mamba-7B-Instruct` on the HateXplain dataset.

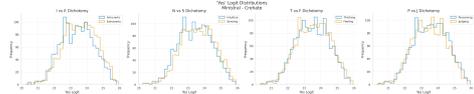

Figure 46: Distribution of yes logits for `Ministral-8B-Instruct` on the CREHate dataset.

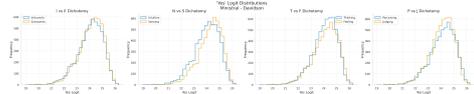

Figure 47: Distribution of yes logits for `Ministral-8B-Instruct` on the Davidson dataset.

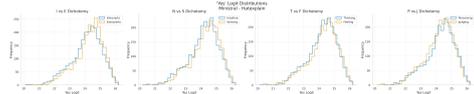

Figure 48: Distribution of yes logits for `Ministral-8B-Instruct` on the HateXplain dataset.

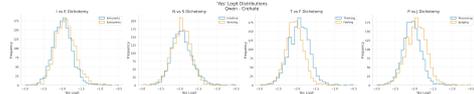

Figure 49: Distribution of yes logits for `Qwen2.5-7B-Instruct` on the CREHate dataset.

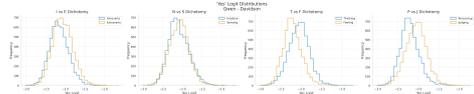

Figure 50: Distribution of yes logits for `Qwen2.5-7B-Instruct` on the Davidson dataset.

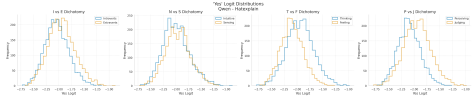

Figure 51: Distribution of yes logits for `Qwen2.5-7B-Instruct` on the HateXplain dataset.

# I Logit Difference Distribution

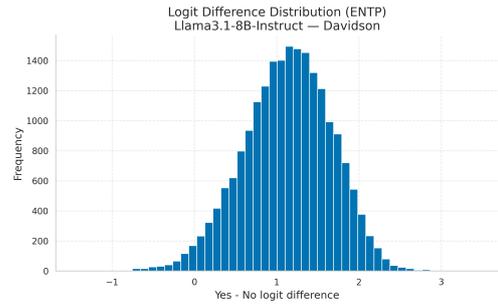

Figure 55: Logit Difference Distribution for `Llama3.1-8B-Instruct` on the Davidson dataset (ENTP).

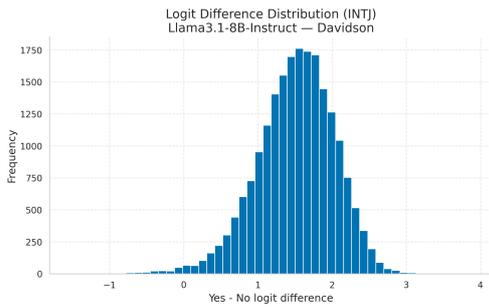

Figure 52: Logit Difference Distribution for `Llama3.1-8B-Instruct` on the Davidson dataset (INTJ).

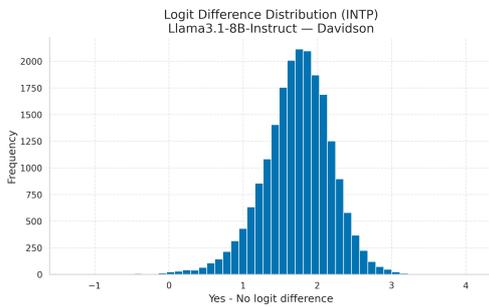

Figure 53: Logit Difference Distribution for `Llama3.1-8B-Instruct` on the Davidson dataset (INTP).

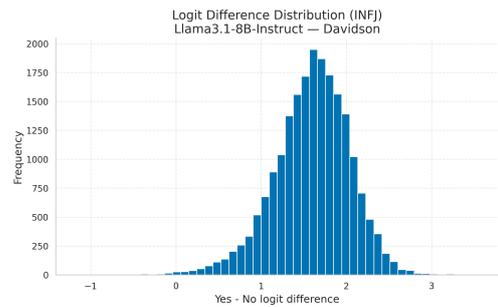

Figure 56: Logit Difference Distribution for `Llama3.1-8B-Instruct` on the Davidson dataset (INFJ).

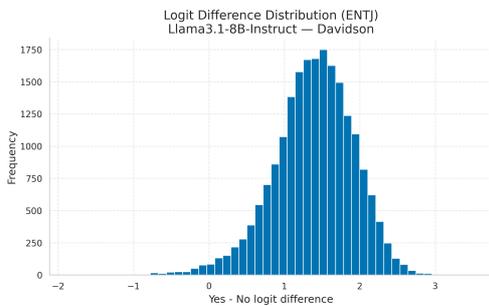

Figure 54: Logit Difference Distribution for `Llama3.1-8B-Instruct` on the Davidson dataset (ENTJ).

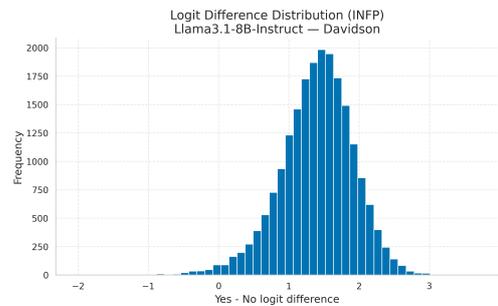

Figure 57: Logit Difference Distribution for `Llama3.1-8B-Instruct` on the Davidson dataset (INFP).

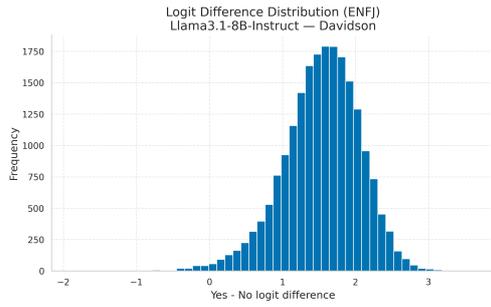

Figure 58: Logit Difference Distribution for `Llama3.1-8B-Instruct` on the Davidson dataset (ENFJ).

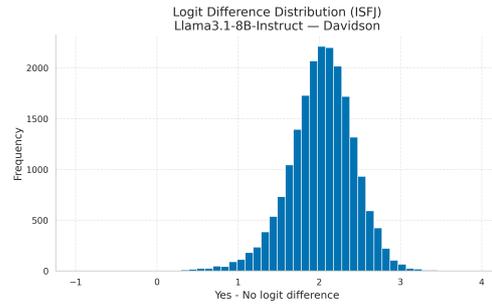

Figure 61: Logit Difference Distribution for `Llama3.1-8B-Instruct` on the Davidson dataset (ISFJ).

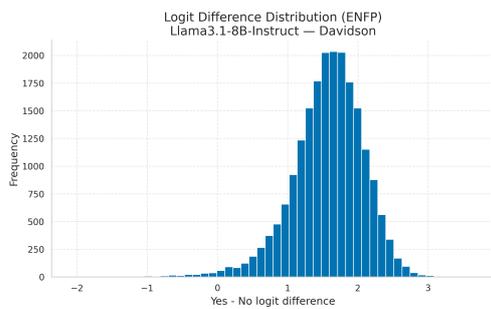

Figure 59: Logit Difference Distribution for `Llama3.1-8B-Instruct` on the Davidson dataset (ENFP).

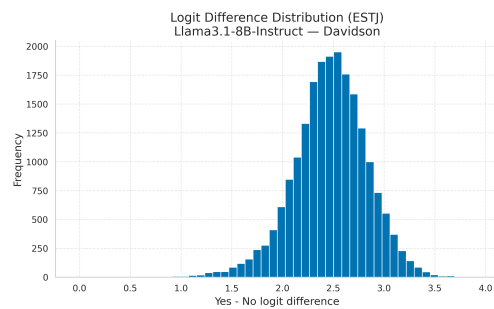

Figure 62: Logit Difference Distribution for `Llama3.1-8B-Instruct` on the Davidson dataset (ESTJ).

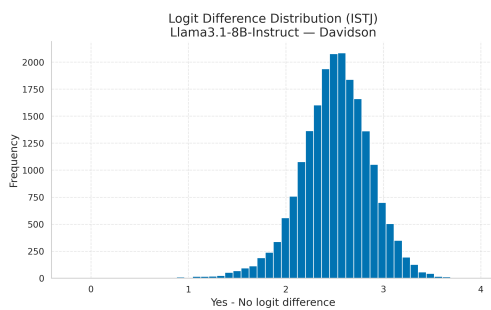

Figure 60: Logit Difference Distribution for `Llama3.1-8B-Instruct` on the Davidson dataset (ISTJ).

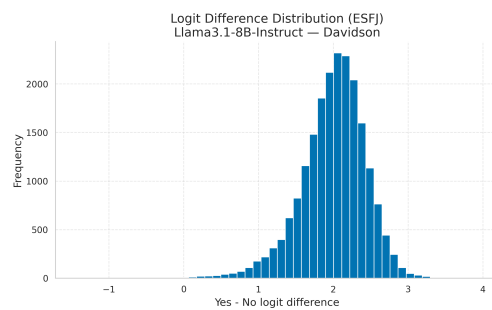

Figure 63: Logit Difference Distribution for `Llama3.1-8B-Instruct` on the Davidson dataset (ESFJ).

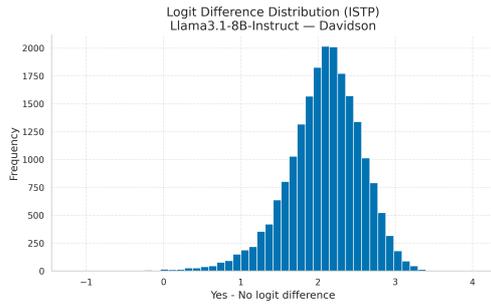

Figure 64: Logit Difference Distribution for `Llama3.1-8B-Instruct` on the Davidson dataset (ISTP).

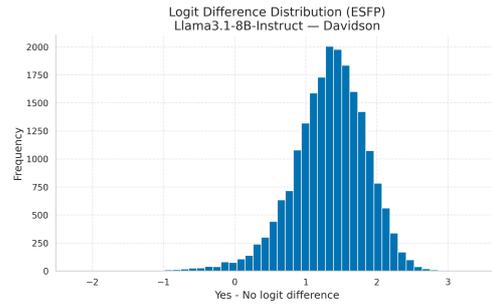

Figure 67: Logit Difference Distribution for `Llama3.1-8B-Instruct` on the Davidson dataset (ESFP).

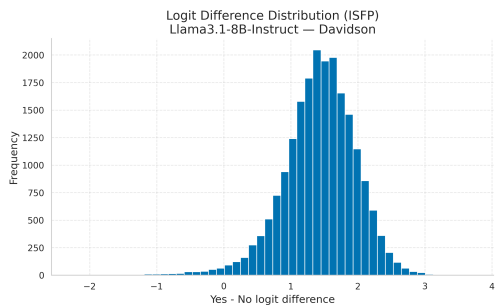

Figure 65: Logit Difference Distribution for `Llama3.1-8B-Instruct` on the Davidson dataset (ISFP).

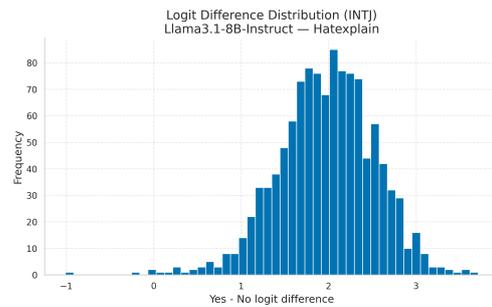

Figure 68: Logit Difference Distribution for `Llama3.1-8B-Instruct` on the Hatexplain dataset (INTJ).

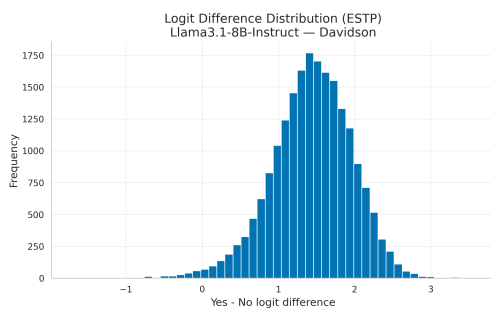

Figure 66: Logit Difference Distribution for `Llama3.1-8B-Instruct` on the Davidson dataset (ESTP).

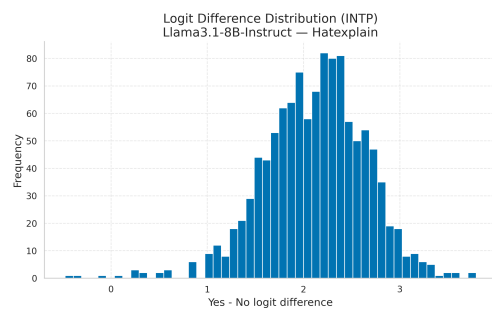

Figure 69: Logit Difference Distribution for `Llama3.1-8B-Instruct` on the Hatexplain dataset (INTP).

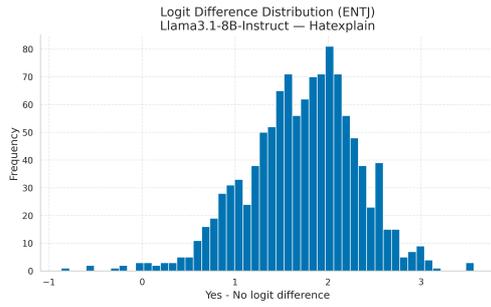

Figure 70: Logit Difference Distribution for `Llama3.1-8B-Instruct` on the Hatexplain dataset (ENTJ).

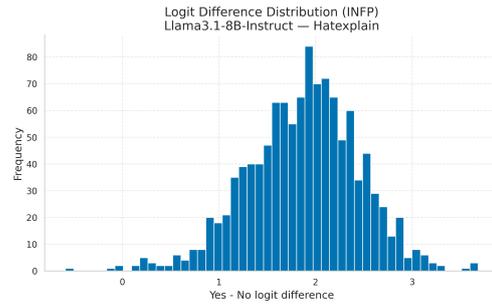

Figure 73: Logit Difference Distribution for `Llama3.1-8B-Instruct` on the Hatexplain dataset (INFP).

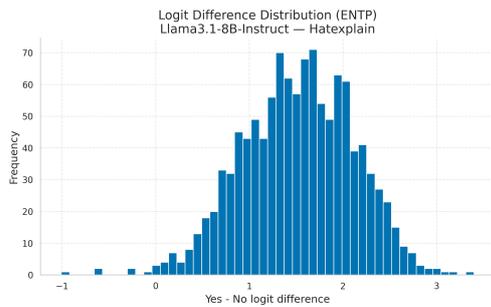

Figure 71: Logit Difference Distribution for `Llama3.1-8B-Instruct` on the Hatexplain dataset (ENTP).

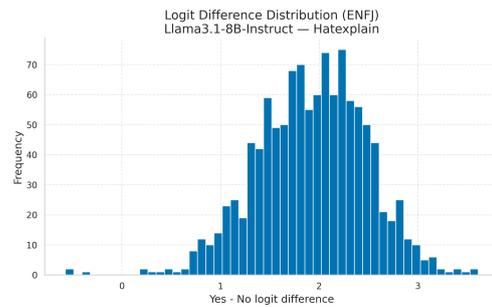

Figure 74: Logit Difference Distribution for `Llama3.1-8B-Instruct` on the Hatexplain dataset (ENFJ).

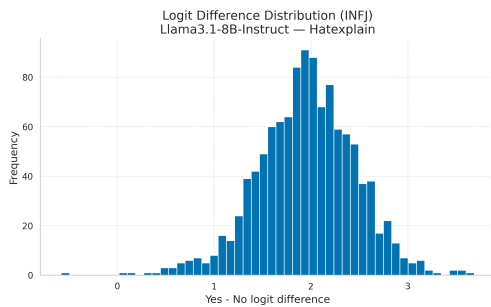

Figure 72: Logit Difference Distribution for `Llama3.1-8B-Instruct` on the Hatexplain dataset (INFJ).

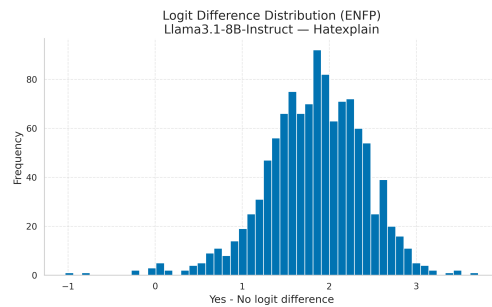

Figure 75: Logit Difference Distribution for `Llama3.1-8B-Instruct` on the Hatexplain dataset (ENFP).

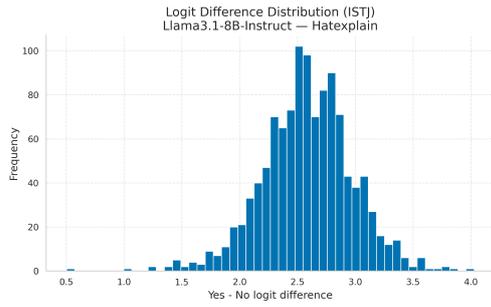

Figure 76: Logit Difference Distribution for `Llama3.1-8B-Instruct` on the Hatexplain dataset (ISTJ).

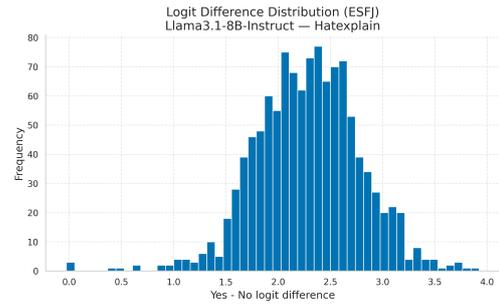

Figure 79: Logit Difference Distribution for `Llama3.1-8B-Instruct` on the Hatexplain dataset (ESFJ).

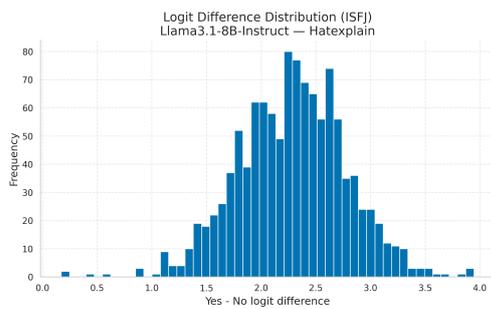

Figure 77: Logit Difference Distribution for `Llama3.1-8B-Instruct` on the Hatexplain dataset (ISFJ).

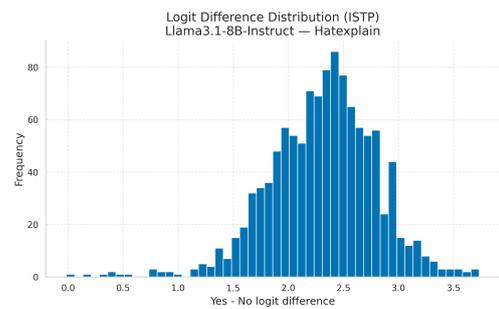

Figure 80: Logit Difference Distribution for `Llama3.1-8B-Instruct` on the Hatexplain dataset (ISTP).

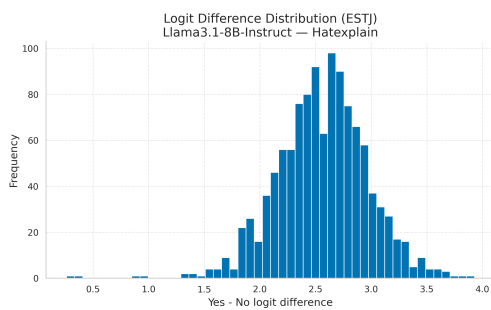

Figure 78: Logit Difference Distribution for `Llama3.1-8B-Instruct` on the Hatexplain dataset (ESTJ).

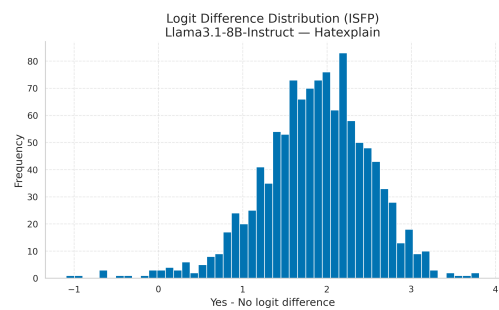

Figure 81: Logit Difference Distribution for `Llama3.1-8B-Instruct` on the Hatexplain dataset (ISFP).

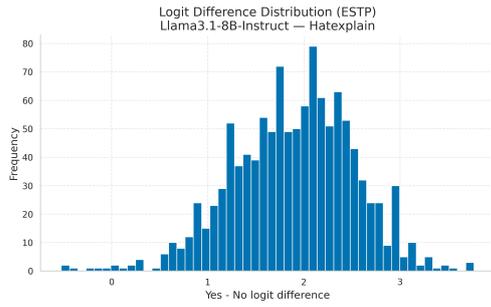

Figure 82: Logit Difference Distribution for `Llama3.1-8B-Instruct` on the Hatexplain dataset (ESTP).

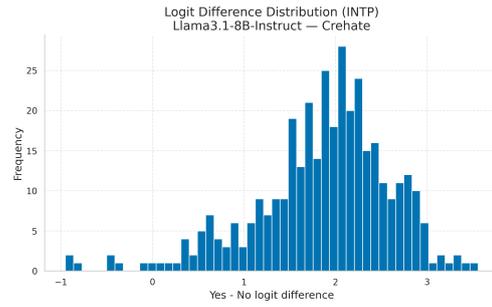

Figure 85: Logit Difference Distribution for `Llama3.1-8B-Instruct` on the Crehate dataset (INTP).

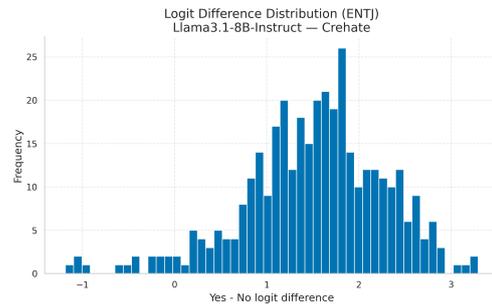

Figure 86: Logit Difference Distribution for `Llama3.1-8B-Instruct` on the Crehate dataset (ENTJ).

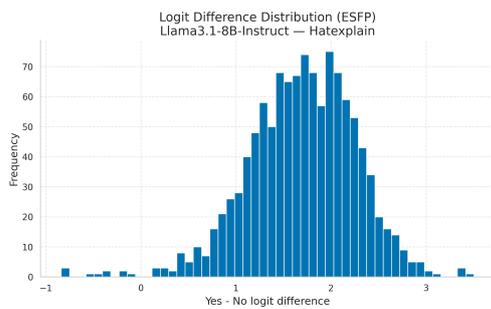

Figure 83: Logit Difference Distribution for `Llama3.1-8B-Instruct` on the Hatexplain dataset (ESFP).

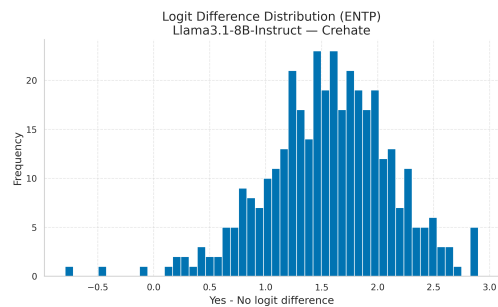

Figure 87: Logit Difference Distribution for `Llama3.1-8B-Instruct` on the Crehate dataset (ENTP).

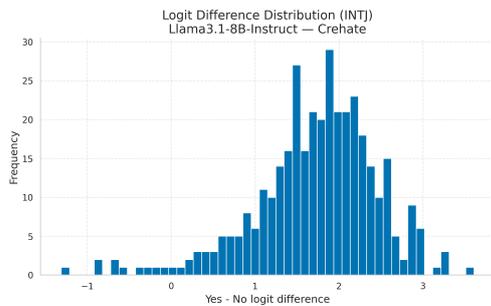

Figure 84: Logit Difference Distribution for `Llama3.1-8B-Instruct` on the Crehate dataset (INTJ).

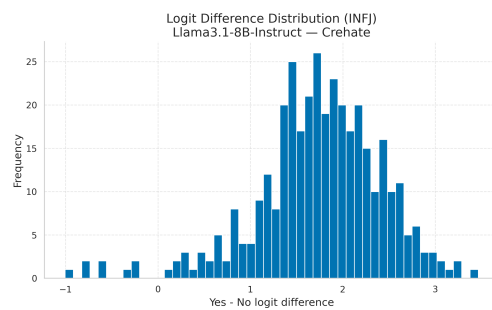

Figure 88: Logit Difference Distribution for `Llama3.1-8B-Instruct` on the Crehate dataset (INFJ).

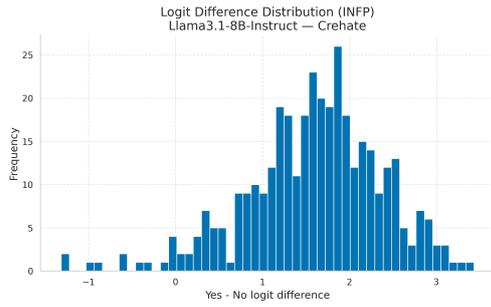

Figure 89: Logit Difference Distribution for `Llama3.1-8B-Instruct` on the Crehate dataset (INFP).

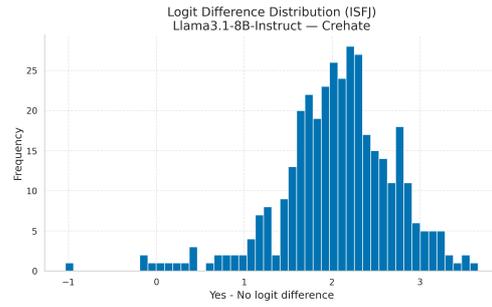

Figure 93: Logit Difference Distribution for `Llama3.1-8B-Instruct` on the Crehate dataset (ISFJ).

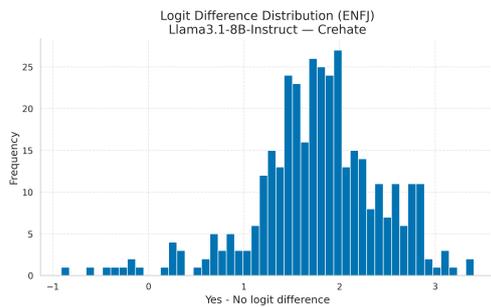

Figure 90: Logit Difference Distribution for `Llama3.1-8B-Instruct` on the Crehate dataset (ENFJ).

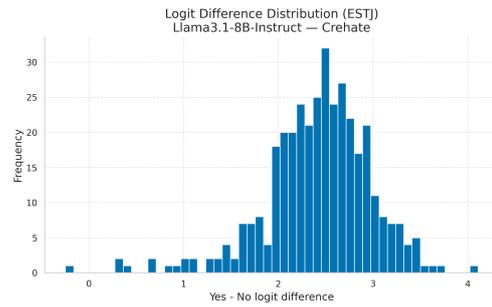

Figure 94: Logit Difference Distribution for `Llama3.1-8B-Instruct` on the Crehate dataset (ESTJ).

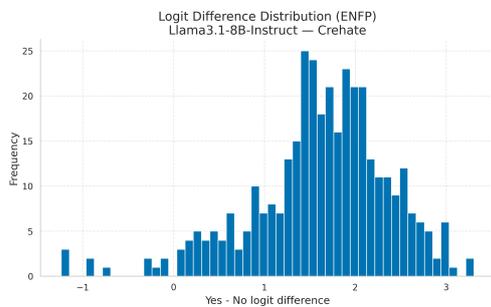

Figure 91: Logit Difference Distribution for `Llama3.1-8B-Instruct` on the Crehate dataset (ENFP).

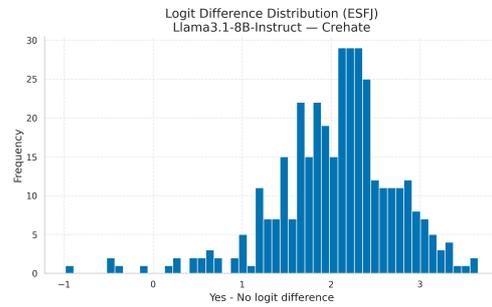

Figure 95: Logit Difference Distribution for `Llama3.1-8B-Instruct` on the Crehate dataset (ESFJ).

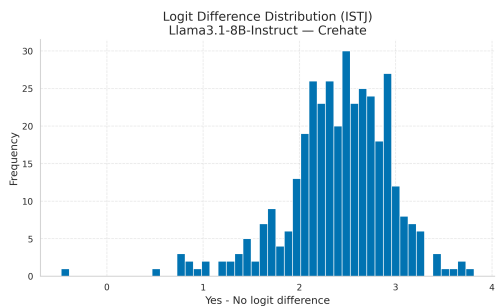

Figure 92: Logit Difference Distribution for `Llama3.1-8B-Instruct` on the Crehate dataset (ISTJ).

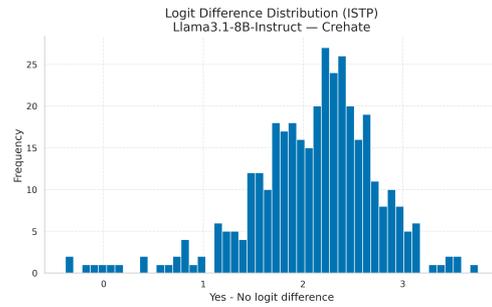

Figure 96: Logit Difference Distribution for `Llama3.1-8B-Instruct` on the Crehate dataset (ISTP).

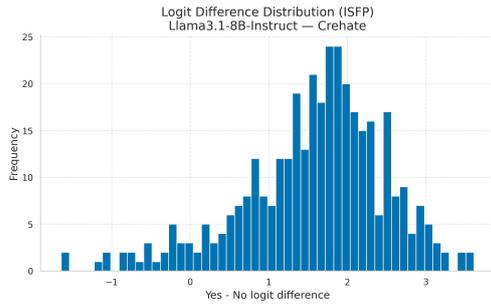

Figure 97: Logit Difference Distribution for `Llama3.1-8B-Instruct` on the Crehate dataset (ISFP).

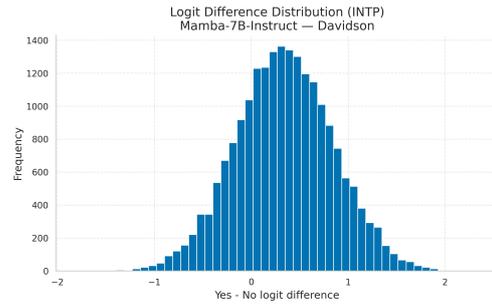

Figure 101: Logit Difference Distribution for `Mamba-7B-Instruct` on the Davidson dataset (INTP).

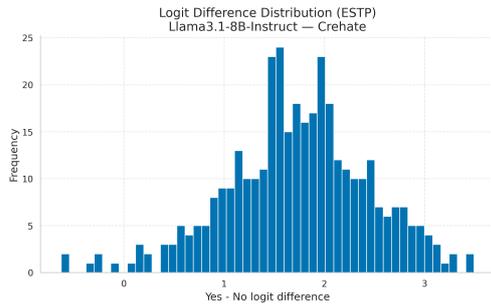

Figure 98: Logit Difference Distribution for `Llama3.1-8B-Instruct` on the Crehate dataset (ESTP).

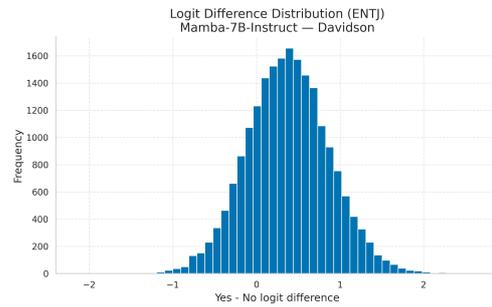

Figure 102: Logit Difference Distribution for `Mamba-7B-Instruct` on the Davidson dataset (ENTJ).

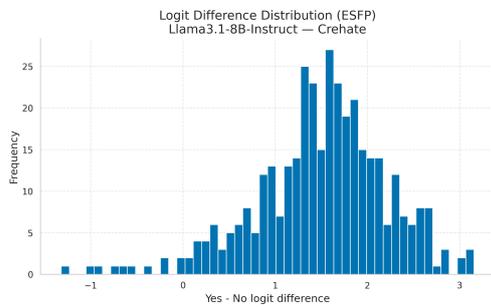

Figure 99: Logit Difference Distribution for `Llama3.1-8B-Instruct` on the Crehate dataset (ESFP).

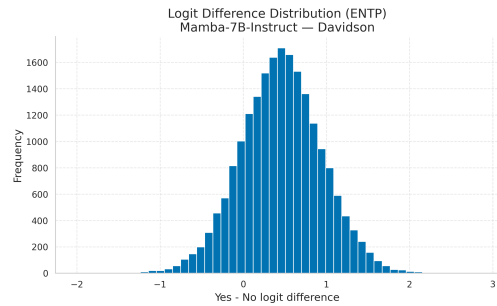

Figure 103: Logit Difference Distribution for `Mamba-7B-Instruct` on the Davidson dataset (ENTP).

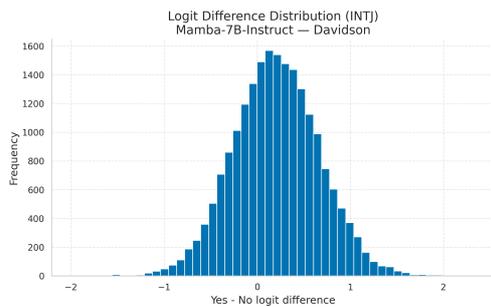

Figure 100: Logit Difference Distribution for `Mamba-7B-Instruct` on the Davidson dataset (INTJ).

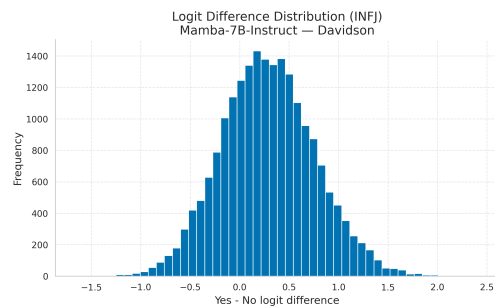

Figure 104: Logit Difference Distribution for `Mamba-7B-Instruct` on the Davidson dataset (INFJ).

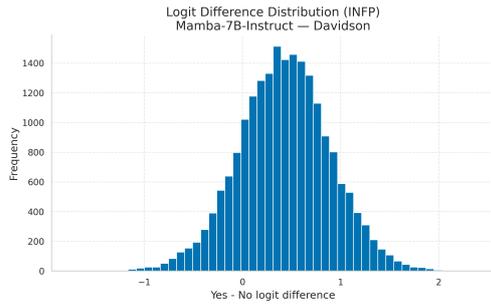

Figure 105: Logit Difference Distribution for `Mamba-7B-Instruct` on the Davidson dataset (INFP).

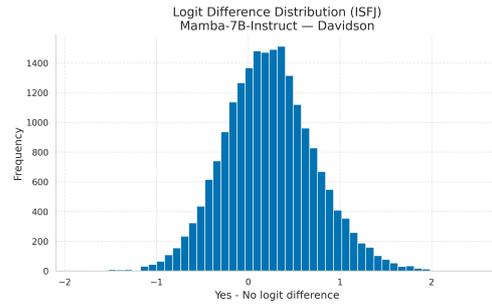

Figure 109: Logit Difference Distribution for `Mamba-7B-Instruct` on the Davidson dataset (ISFJ).

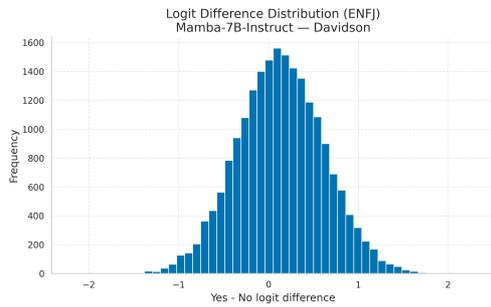

Figure 106: Logit Difference Distribution for `Mamba-7B-Instruct` on the Davidson dataset (ENFJ).

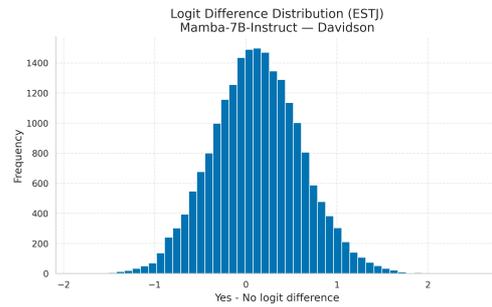

Figure 110: Logit Difference Distribution for `Mamba-7B-Instruct` on the Davidson dataset (ESTJ).

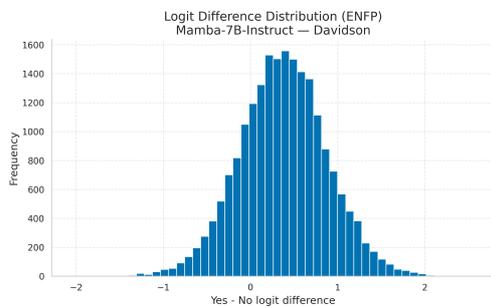

Figure 107: Logit Difference Distribution for `Mamba-7B-Instruct` on the Davidson dataset (ENFP).

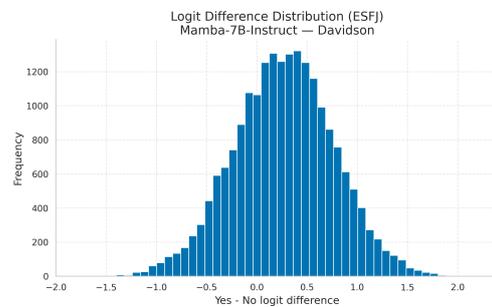

Figure 111: Logit Difference Distribution for `Mamba-7B-Instruct` on the Davidson dataset (ESFJ).

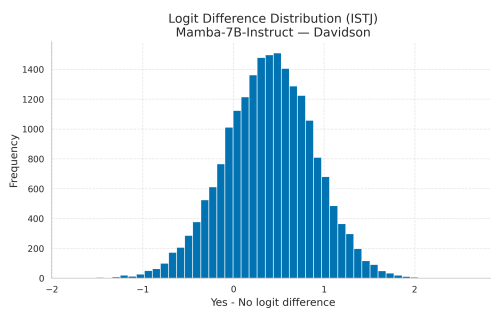

Figure 108: Logit Difference Distribution for `Mamba-7B-Instruct` on the Davidson dataset (ISTJ).

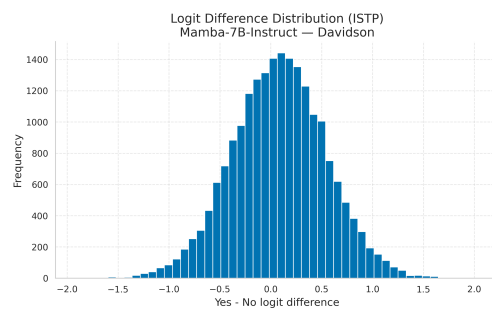

Figure 112: Logit Difference Distribution for `Mamba-7B-Instruct` on the Davidson dataset (ISTP).

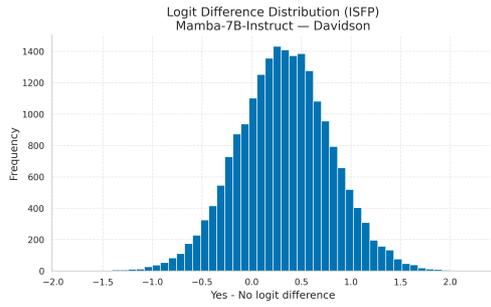

Figure 113: Logit Difference Distribution for `Mamba-7B-Instruct` on the Davidson dataset (ISFP).

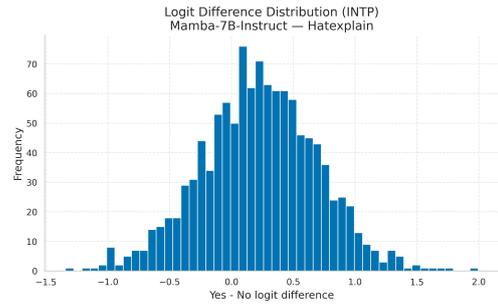

Figure 117: Logit Difference Distribution for `Mamba-7B-Instruct` on the Hatexplain dataset (INTP).

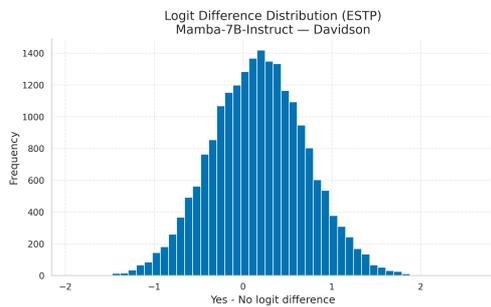

Figure 114: Logit Difference Distribution for `Mamba-7B-Instruct` on the Davidson dataset (ESTP).

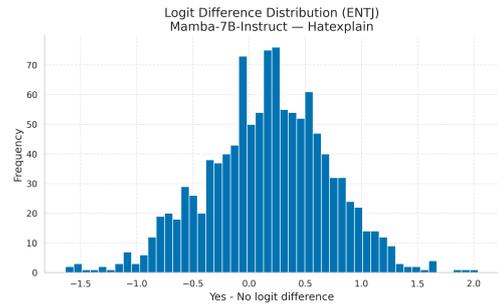

Figure 118: Logit Difference Distribution for `Mamba-7B-Instruct` on the Hatexplain dataset (ENTJ).

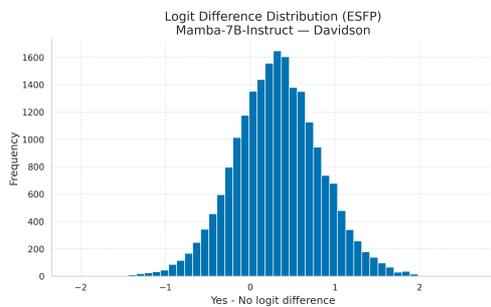

Figure 115: Logit Difference Distribution for `Mamba-7B-Instruct` on the Davidson dataset (ESFP).

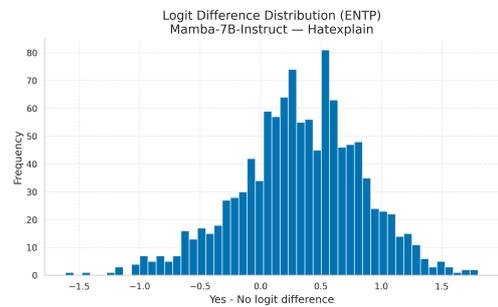

Figure 119: Logit Difference Distribution for `Mamba-7B-Instruct` on the Hatexplain dataset (ENTP).

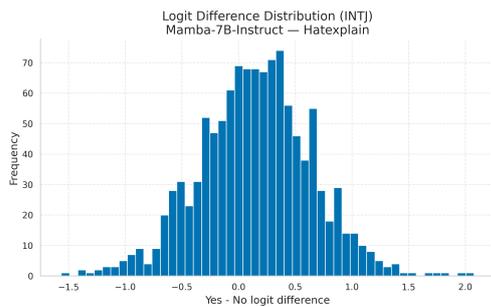

Figure 116: Logit Difference Distribution for `Mamba-7B-Instruct` on the Hatexplain dataset (INTJ).

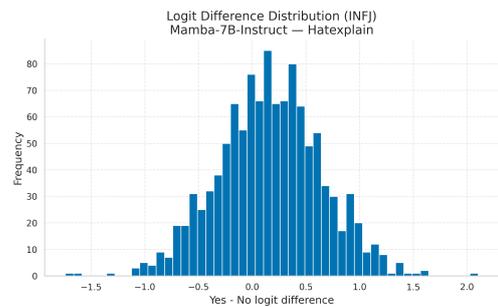

Figure 120: Logit Difference Distribution for `Mamba-7B-Instruct` on the Hatexplain dataset (INFJ).

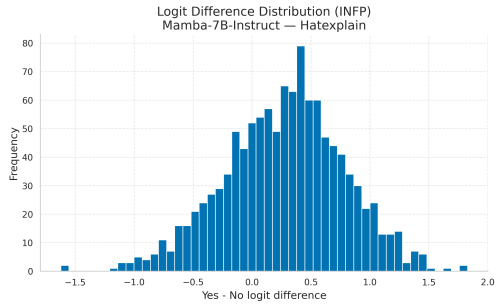

Figure 121: Logit Difference Distribution for `Mamba-7B-Instruct` on the Hatexplain dataset (INFP).

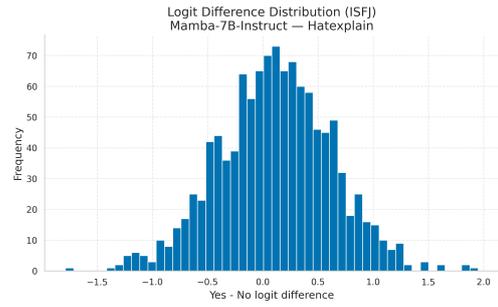

Figure 125: Logit Difference Distribution for `Mamba-7B-Instruct` on the Hatexplain dataset (ISFJ).

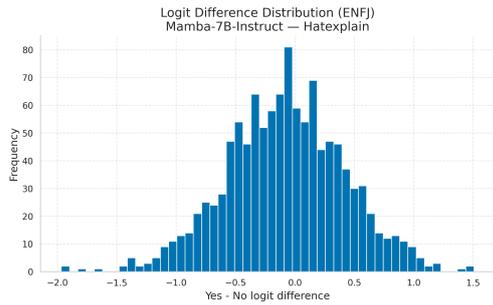

Figure 122: Logit Difference Distribution for `Mamba-7B-Instruct` on the Hatexplain dataset (ENFJ).

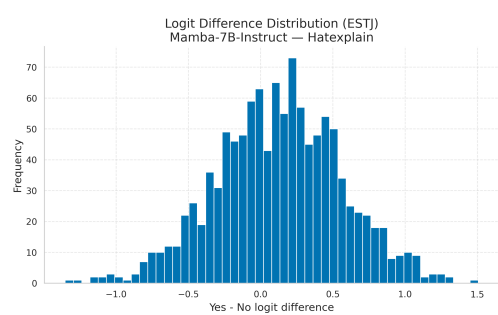

Figure 126: Logit Difference Distribution for `Mamba-7B-Instruct` on the Hatexplain dataset (ESTJ).

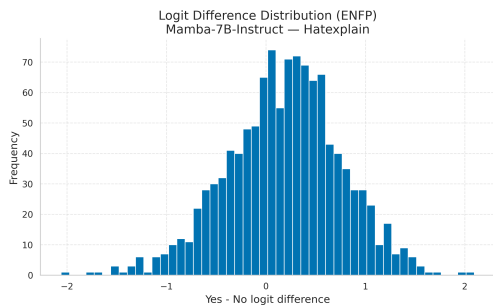

Figure 123: Logit Difference Distribution for `Mamba-7B-Instruct` on the Hatexplain dataset (ENFP).

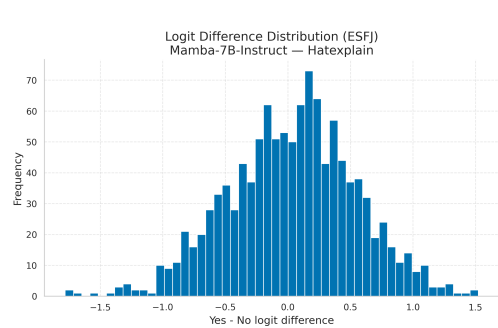

Figure 127: Logit Difference Distribution for `Mamba-7B-Instruct` on the Hatexplain dataset (ESFJ).

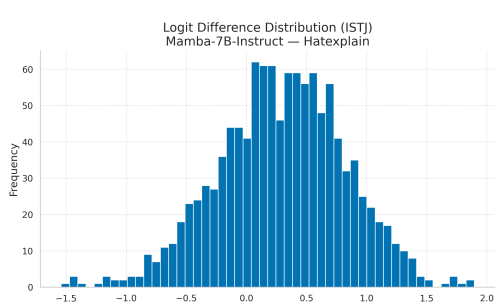

Figure 124: Logit Difference Distribution for `Mamba-7B-Instruct` on the Hatexplain dataset (ISTJ).

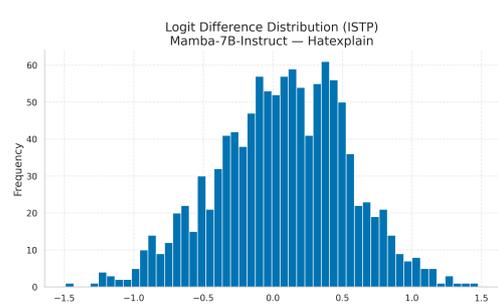

Figure 128: Logit Difference Distribution for `Mamba-7B-Instruct` on the Hatexplain dataset (ISTP).

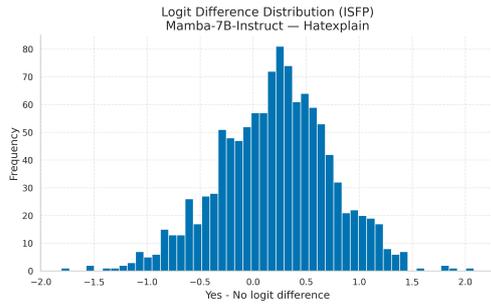

Figure 129: Logit Difference Distribution for Mamba-7B-Instruct on the Hatexplain dataset (ISFP).

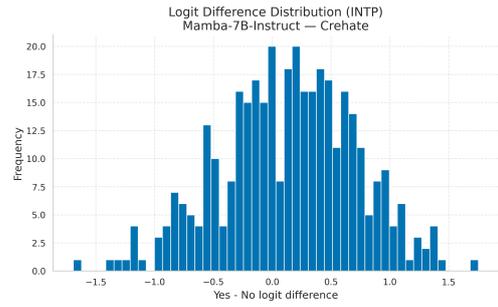

Figure 133: Logit Difference Distribution for Mamba-7B-Instruct on the Crehate dataset (INTP).

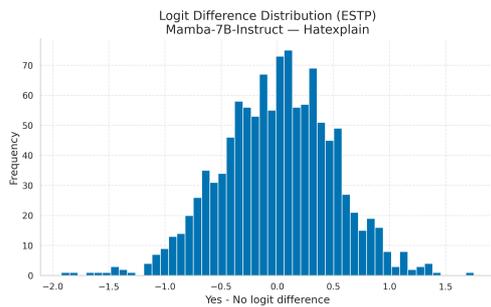

Figure 130: Logit Difference Distribution for Mamba-7B-Instruct on the Hatexplain dataset (ESTP).

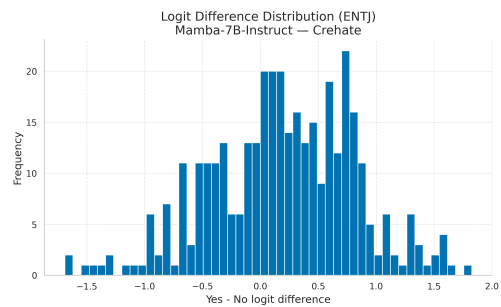

Figure 134: Logit Difference Distribution for Mamba-7B-Instruct on the Crehate dataset (ENTJ).

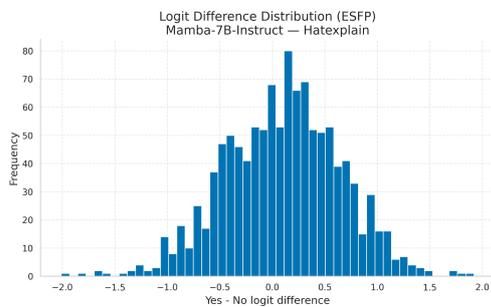

Figure 131: Logit Difference Distribution for Mamba-7B-Instruct on the Hatexplain dataset (ESFP).

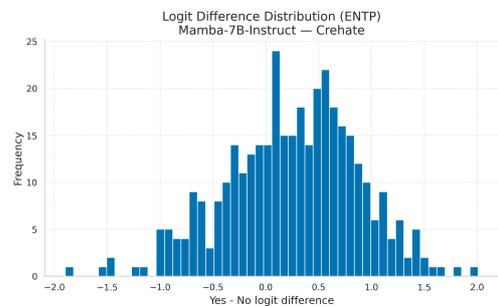

Figure 135: Logit Difference Distribution for Mamba-7B-Instruct on the Crehate dataset (ENTP).

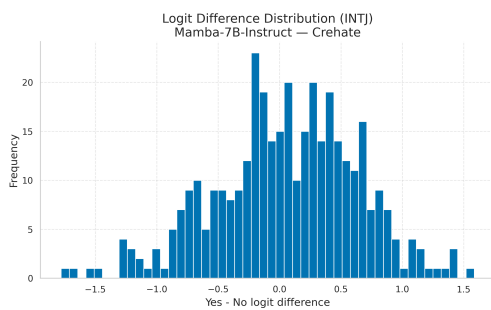

Figure 132: Logit Difference Distribution for Mamba-7B-Instruct on the Crehate dataset (INTJ).

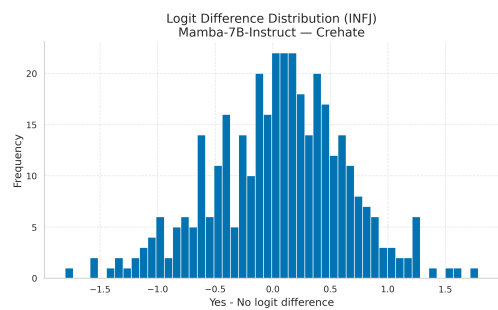

Figure 136: Logit Difference Distribution for Mamba-7B-Instruct on the Crehate dataset (INFJ).

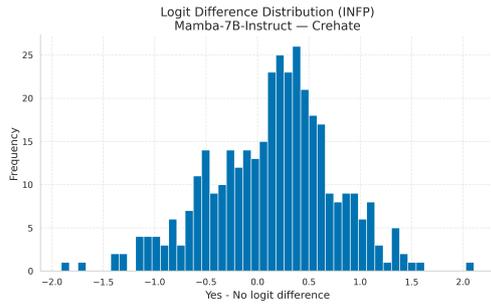

Figure 137: Logit Difference Distribution for `Mamba-7B-Instruct` on the Crehate dataset (INFP).

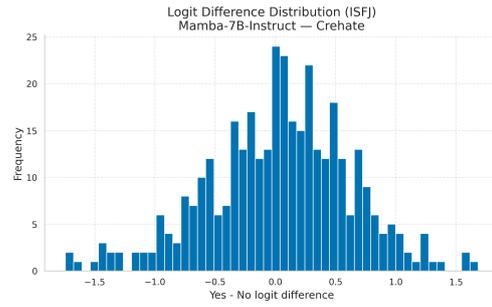

Figure 141: Logit Difference Distribution for `Mamba-7B-Instruct` on the Crehate dataset (ISFJ).

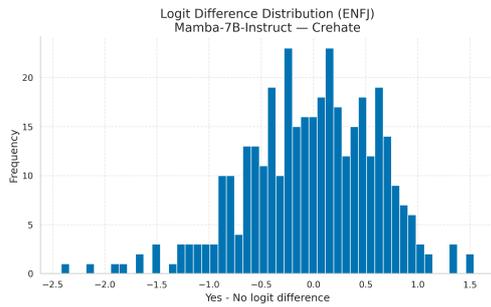

Figure 138: Logit Difference Distribution for `Mamba-7B-Instruct` on the Crehate dataset (ENFJ).

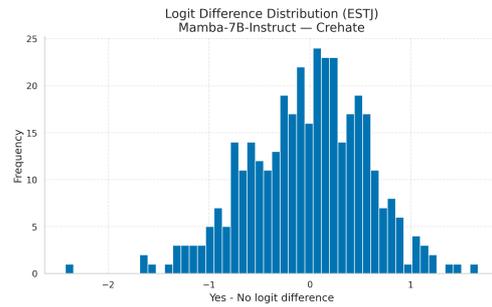

Figure 142: Logit Difference Distribution for `Mamba-7B-Instruct` on the Crehate dataset (ESTJ).

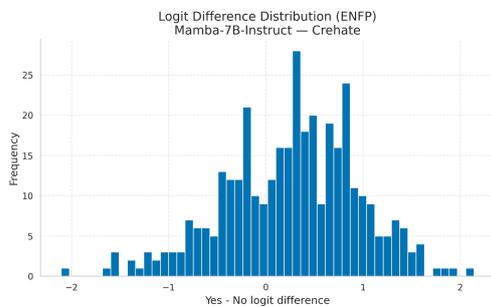

Figure 139: Logit Difference Distribution for `Mamba-7B-Instruct` on the Crehate dataset (ENFP).

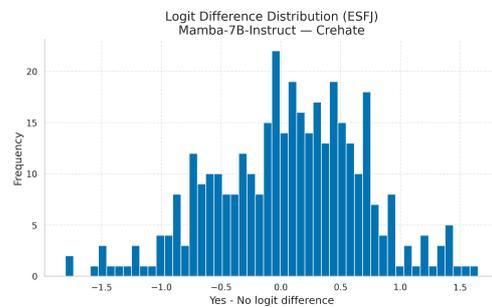

Figure 143: Logit Difference Distribution for `Mamba-7B-Instruct` on the Crehate dataset (ESFJ).

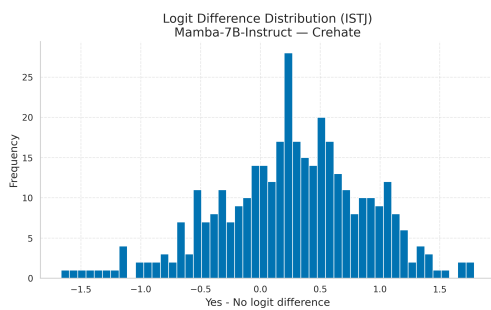

Figure 140: Logit Difference Distribution for `Mamba-7B-Instruct` on the Crehate dataset (ISTJ).

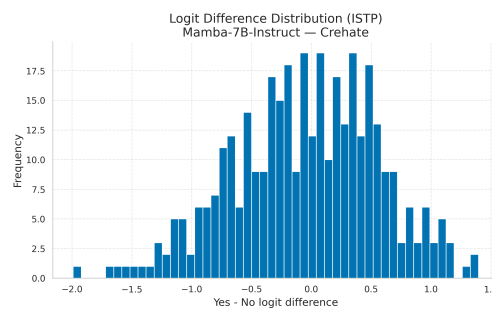

Figure 144: Logit Difference Distribution for `Mamba-7B-Instruct` on the Crehate dataset (ISTP).

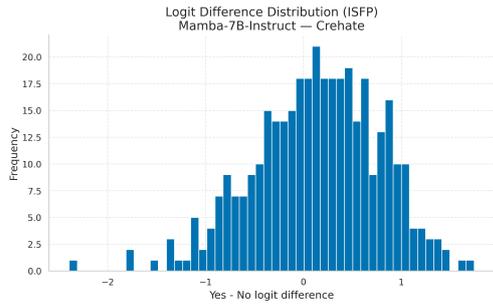

Figure 145: Logit Difference Distribution for `Mamba-7B-Instruct` on the Crehate dataset (ISFP).

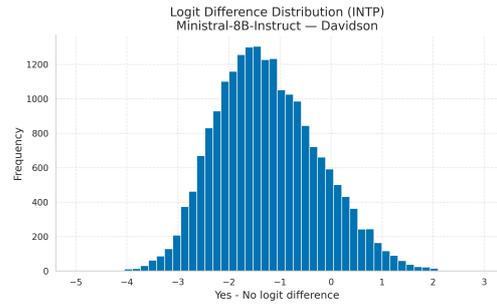

Figure 149: Logit Difference Distribution for `Ministral-8B-Instruct` on the Davidson dataset (INTP).

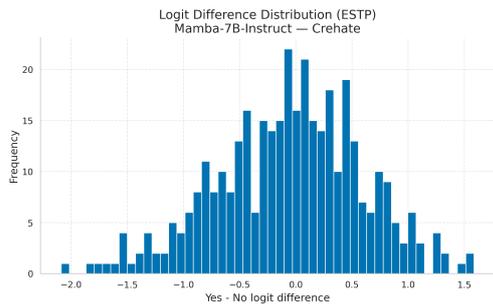

Figure 146: Logit Difference Distribution for `Mamba-7B-Instruct` on the Crehate dataset (ESTP).

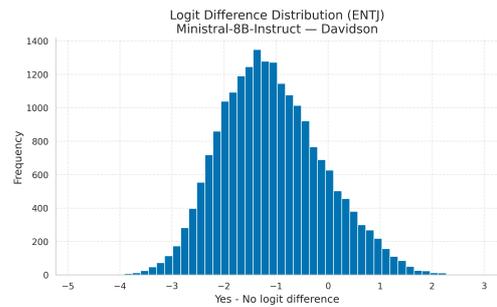

Figure 150: Logit Difference Distribution for `Ministral-8B-Instruct` on the Davidson dataset (ENTJ).

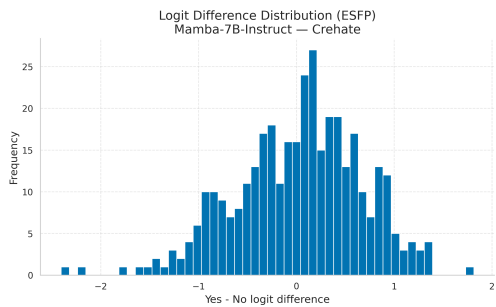

Figure 147: Logit Difference Distribution for `Mamba-7B-Instruct` on the Crehate dataset (ESFP).

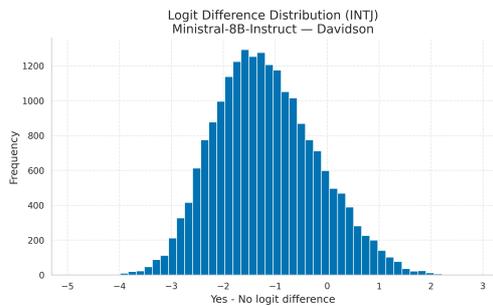

Figure 148: Logit Difference Distribution for `Ministral-8B-Instruct` on the Davidson dataset (INTJ).

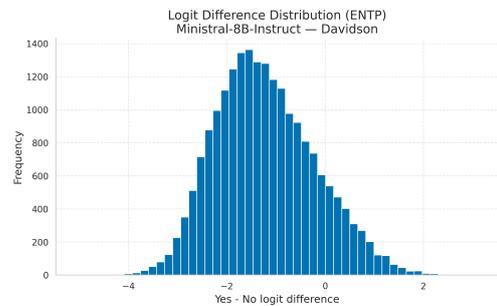

Figure 151: Logit Difference Distribution for `Ministral-8B-Instruct` on the Davidson dataset (ENTP).

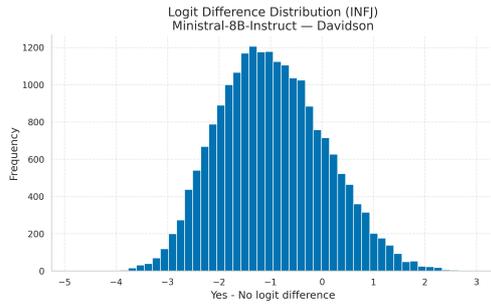

Figure 152: Logit Difference Distribution for `Ministral-8B-Instruct` on the Davidson dataset (INFJ).

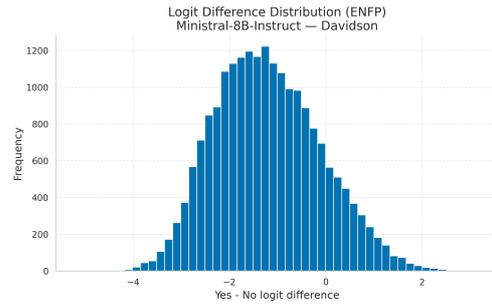

Figure 155: Logit Difference Distribution for `Ministral-8B-Instruct` on the Davidson dataset (ENFP).

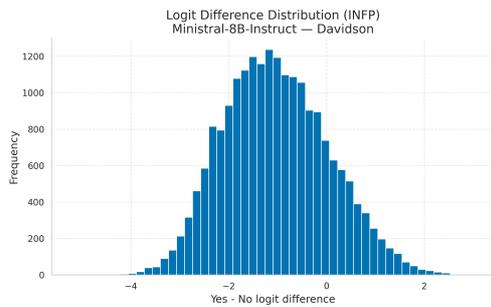

Figure 153: Logit Difference Distribution for `Ministral-8B-Instruct` on the Davidson dataset (INFP).

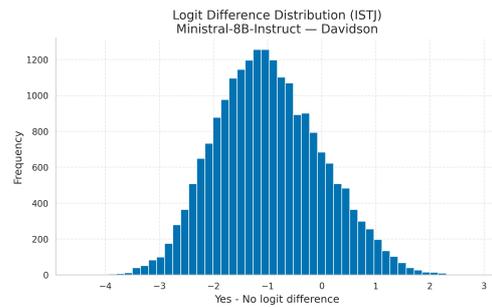

Figure 156: Logit Difference Distribution for `Ministral-8B-Instruct` on the Davidson dataset (ISTJ).

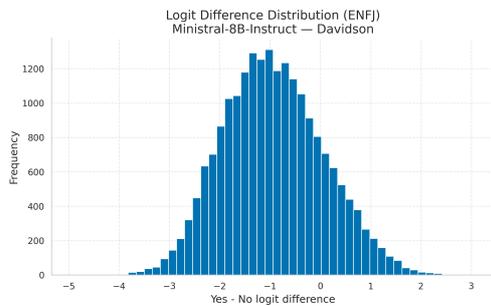

Figure 154: Logit Difference Distribution for `Ministral-8B-Instruct` on the Davidson dataset (ENFJ).

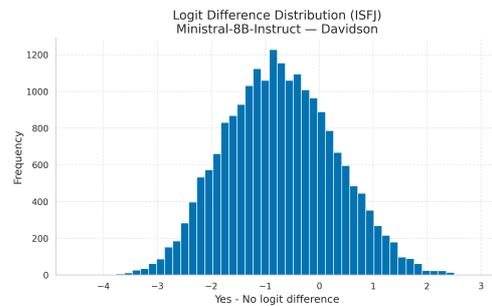

Figure 157: Logit Difference Distribution for `Ministral-8B-Instruct` on the Davidson dataset (ISFJ).

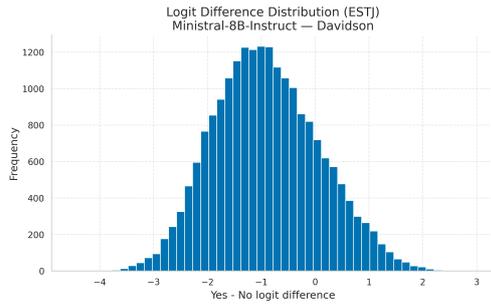

Figure 158: Logit Difference Distribution for `Ministral-8B-Instruct` on the Davidson dataset (ESTJ).

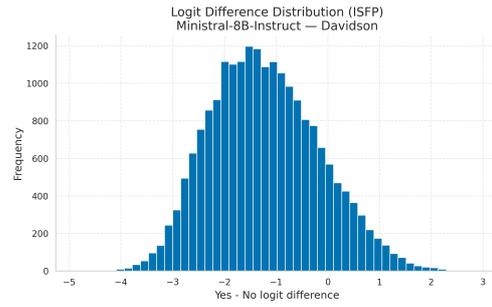

Figure 161: Logit Difference Distribution for `Ministral-8B-Instruct` on the Davidson dataset (ISFP).

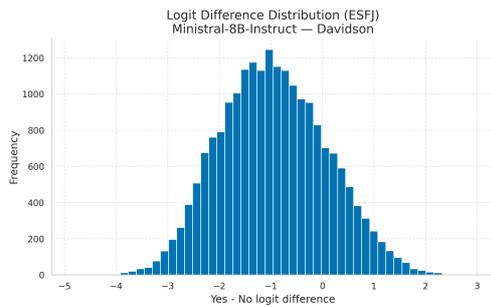

Figure 159: Logit Difference Distribution for `Ministral-8B-Instruct` on the Davidson dataset (ESFJ).

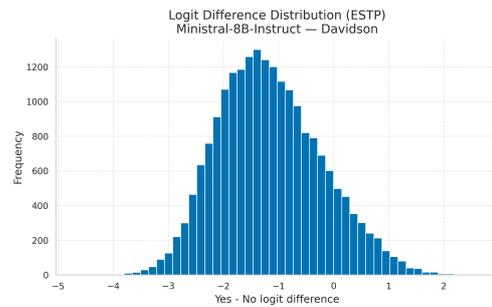

Figure 162: Logit Difference Distribution for `Ministral-8B-Instruct` on the Davidson dataset (ESTP).

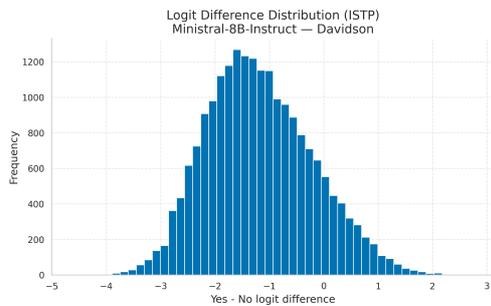

Figure 160: Logit Difference Distribution for `Ministral-8B-Instruct` on the Davidson dataset (ISTP).

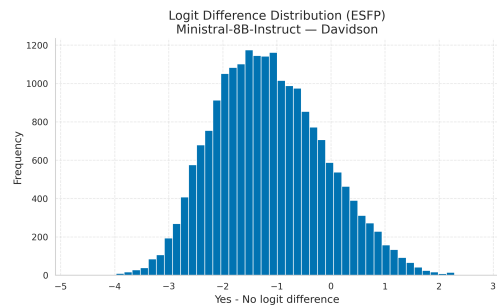

Figure 163: Logit Difference Distribution for `Ministral-8B-Instruct` on the Davidson dataset (ESFP).

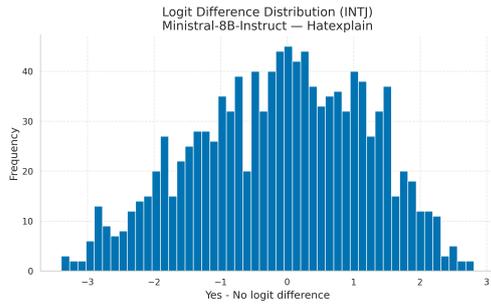

Figure 164: Logit Difference Distribution for `Ministral-8B-Instruct` on the Hatexplain dataset (INTJ).

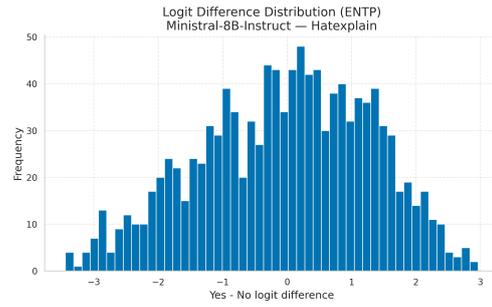

Figure 167: Logit Difference Distribution for `Ministral-8B-Instruct` on the Hatexplain dataset (ENTP).

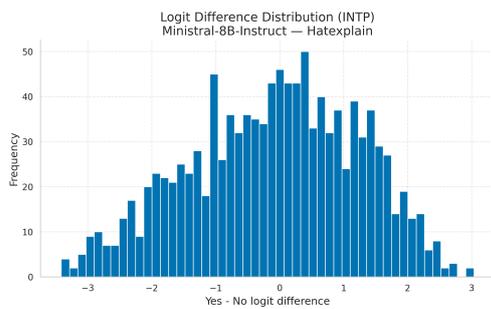

Figure 165: Logit Difference Distribution for `Ministral-8B-Instruct` on the Hatexplain dataset (INTP).

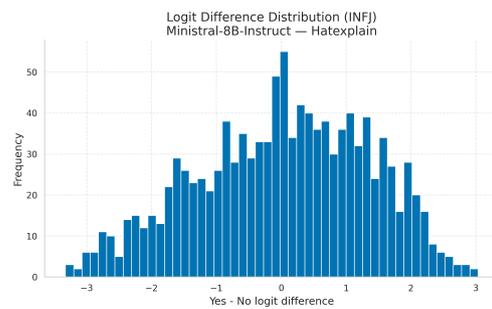

Figure 168: Logit Difference Distribution for `Ministral-8B-Instruct` on the Hatexplain dataset (INFJ).

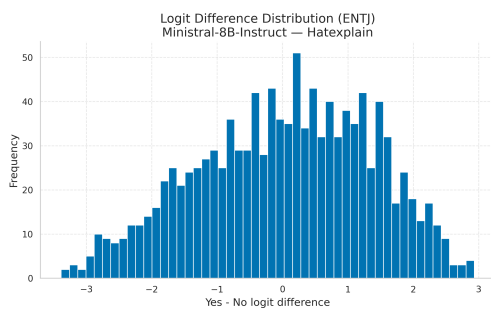

Figure 166: Logit Difference Distribution for `Ministral-8B-Instruct` on the Hatexplain dataset (ENTJ).

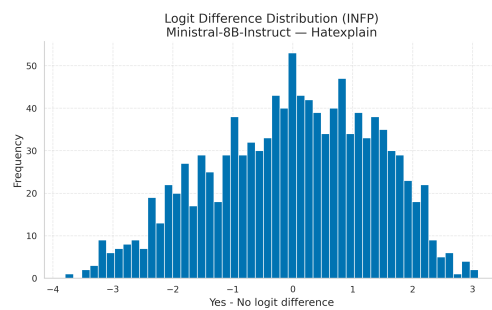

Figure 169: Logit Difference Distribution for `Ministral-8B-Instruct` on the Hatexplain dataset (INFP).

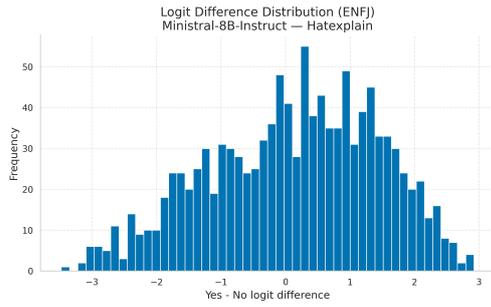

Figure 170: Logit Difference Distribution for `Ministral-8B-Instruct` on the Hatexplain dataset (ENFJ).

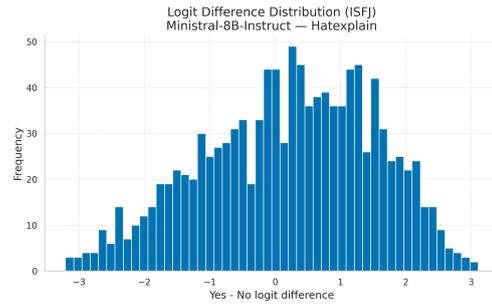

Figure 173: Logit Difference Distribution for `Ministral-8B-Instruct` on the Hatexplain dataset (ISFJ).

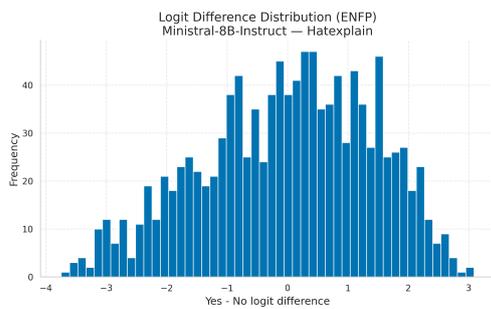

Figure 171: Logit Difference Distribution for `Ministral-8B-Instruct` on the Hatexplain dataset (ENFP).

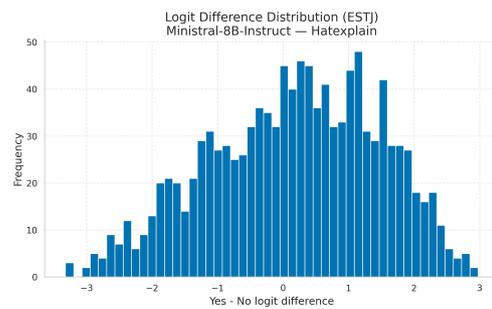

Figure 174: Logit Difference Distribution for `Ministral-8B-Instruct` on the Hatexplain dataset (ESTJ).

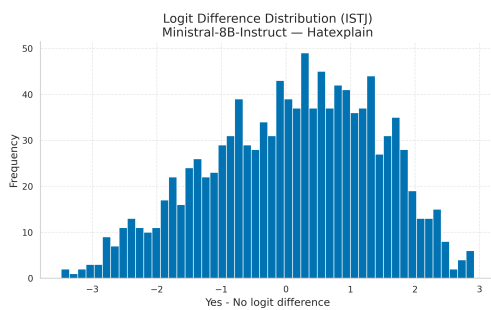

Figure 172: Logit Difference Distribution for `Ministral-8B-Instruct` on the Hatexplain dataset (ISTJ).

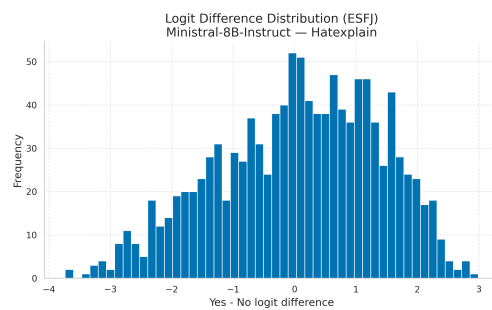

Figure 175: Logit Difference Distribution for `Ministral-8B-Instruct` on the Hatexplain dataset (ESFJ).

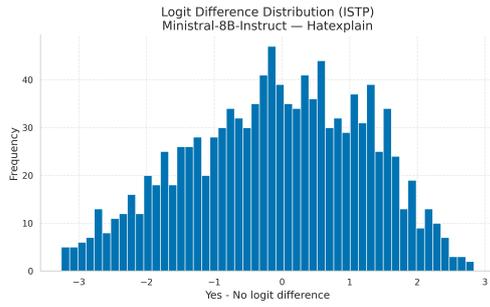

Figure 176: Logit Difference Distribution for `Ministral-8B-Instruct` on the Hatexplain dataset (ISTP).

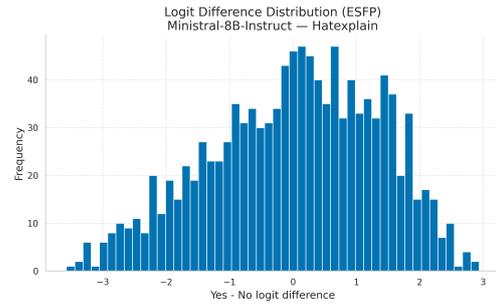

Figure 179: Logit Difference Distribution for `Ministral-8B-Instruct` on the Hatexplain dataset (ESFP).

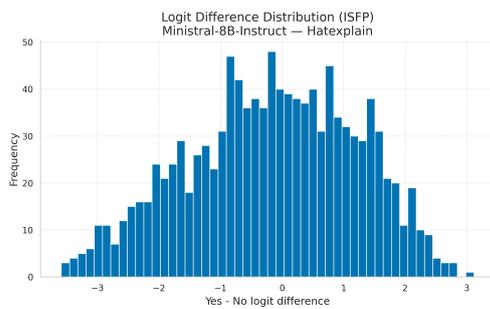

Figure 177: Logit Difference Distribution for `Ministral-8B-Instruct` on the Hatexplain dataset (ISFP).

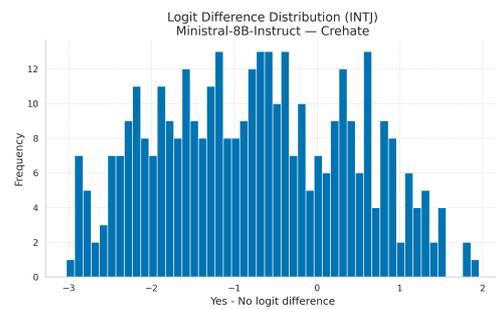

Figure 180: Logit Difference Distribution for `Ministral-8B-Instruct` on the Crehate dataset (INTJ).

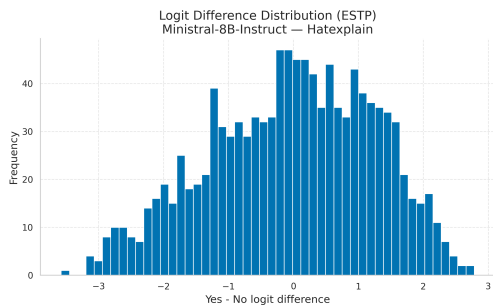

Figure 178: Logit Difference Distribution for `Ministral-8B-Instruct` on the Hatexplain dataset (ESTP).

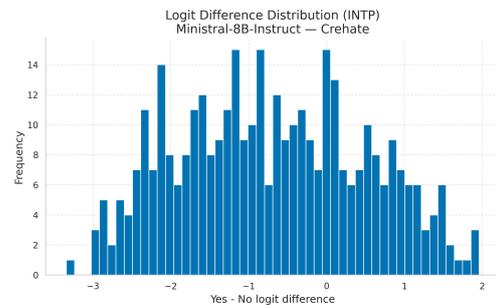

Figure 181: Logit Difference Distribution for `Ministral-8B-Instruct` on the Crehate dataset (INTP).

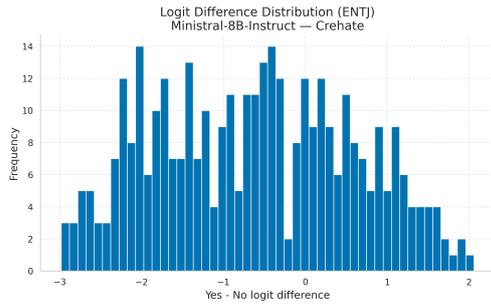

Figure 182: Logit Difference Distribution for `Ministral-8B-Instruct` on the Crehate dataset (ENTJ).

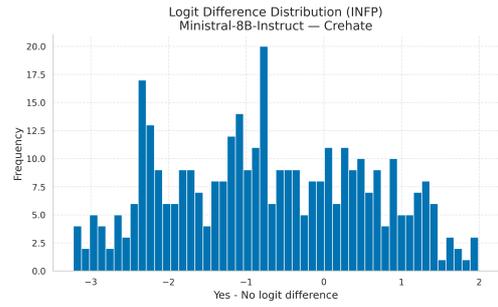

Figure 185: Logit Difference Distribution for `Ministral-8B-Instruct` on the Crehate dataset (INFP).

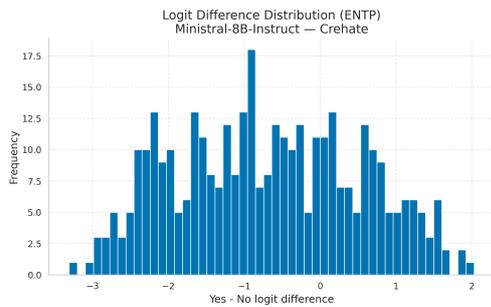

Figure 183: Logit Difference Distribution for `Ministral-8B-Instruct` on the Crehate dataset (ENTP).

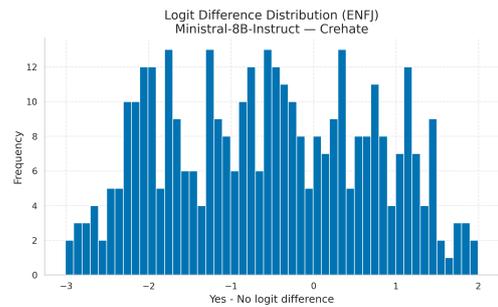

Figure 186: Logit Difference Distribution for `Ministral-8B-Instruct` on the Crehate dataset (ENFJ).

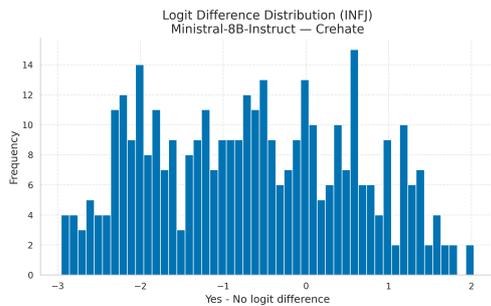

Figure 184: Logit Difference Distribution for `Ministral-8B-Instruct` on the Crehate dataset (INFJ).

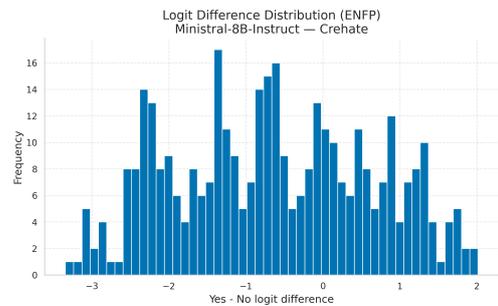

Figure 187: Logit Difference Distribution for `Ministral-8B-Instruct` on the Crehate dataset (ENFP).

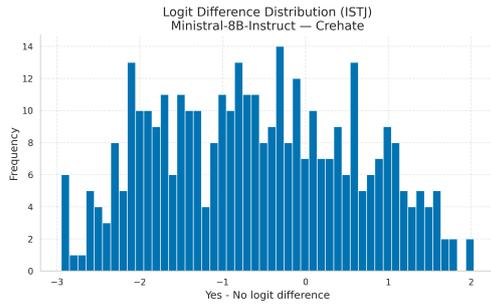

Figure 188: Logit Difference Distribution for `Ministral-8B-Instruct` on the Crehate dataset (ISTJ).

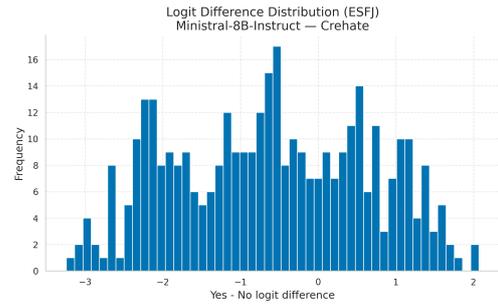

Figure 191: Logit Difference Distribution for `Ministral-8B-Instruct` on the Crehate dataset (ESFJ).

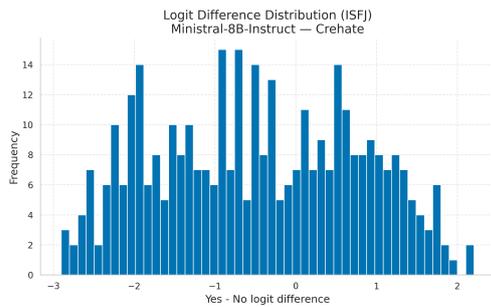

Figure 189: Logit Difference Distribution for `Ministral-8B-Instruct` on the Crehate dataset (ISFJ).

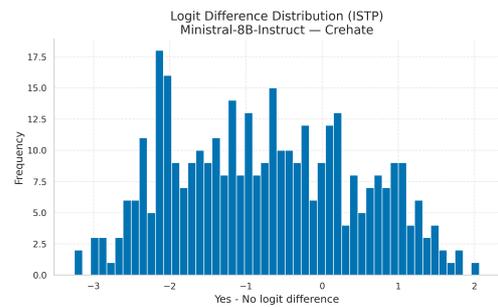

Figure 192: Logit Difference Distribution for `Ministral-8B-Instruct` on the Crehate dataset (ISTP).

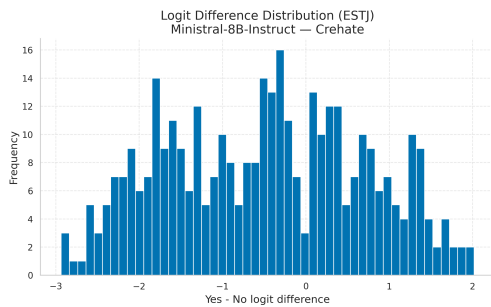

Figure 190: Logit Difference Distribution for `Ministral-8B-Instruct` on the Crehate dataset (ESTJ).

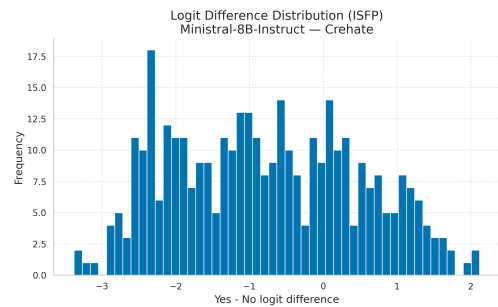

Figure 193: Logit Difference Distribution for `Ministral-8B-Instruct` on the Crehate dataset (ISFP).

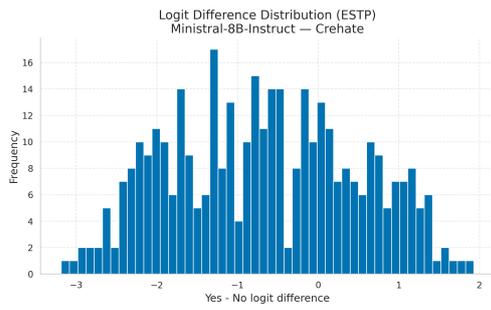

Figure 194: Logit Difference Distribution for `Ministral-8B-Instruct` on the Crehate dataset (ESTP).

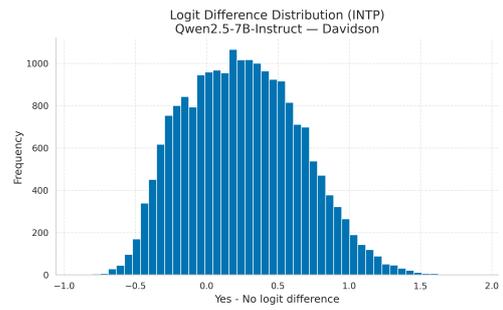

Figure 197: Logit Difference Distribution for `Qwen2.5-7B-Instruct` on the Davidson dataset (INTP).

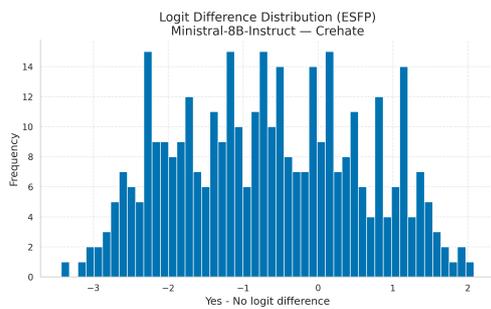

Figure 195: Logit Difference Distribution for `Ministral-8B-Instruct` on the Crehate dataset (ESFP).

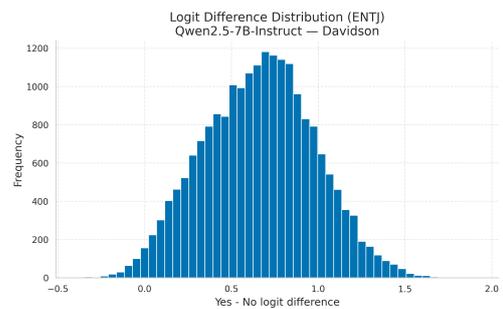

Figure 198: Logit Difference Distribution for `Qwen2.5-7B-Instruct` on the Davidson dataset (ENTJ).

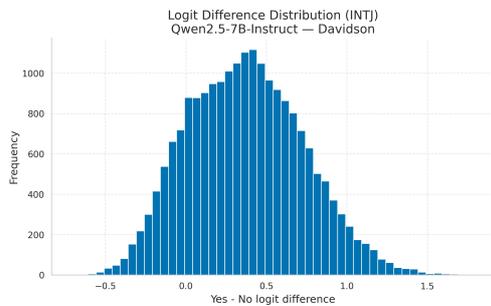

Figure 196: Logit Difference Distribution for `Qwen2.5-7B-Instruct` on the Davidson dataset (INTJ).

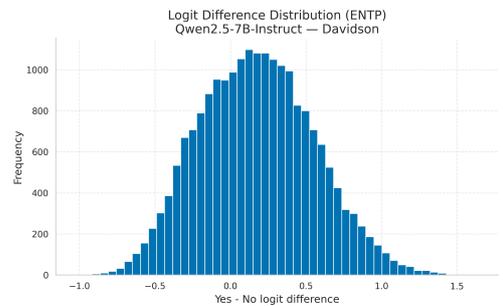

Figure 199: Logit Difference Distribution for `Qwen2.5-7B-Instruct` on the Davidson dataset (ENTP).

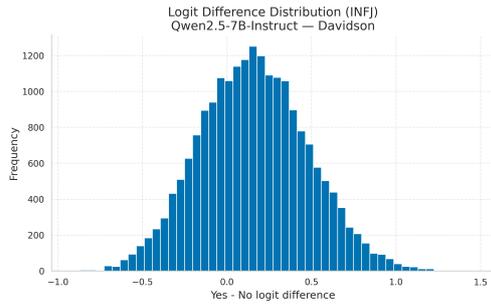

Figure 200: Logit Difference Distribution for `Qwen2.5-7B-Instruct` on the Davidson dataset (INFJ).

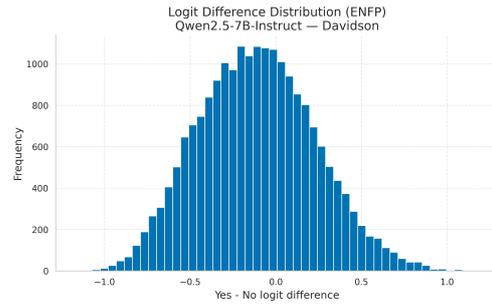

Figure 203: Logit Difference Distribution for `Qwen2.5-7B-Instruct` on the Davidson dataset (ENFP).

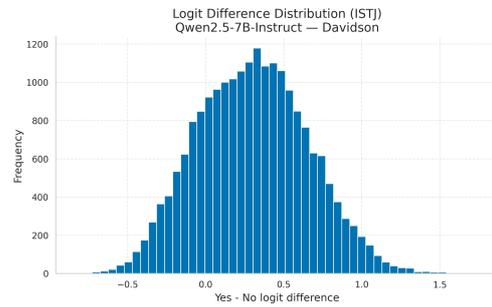

Figure 204: Logit Difference Distribution for `Qwen2.5-7B-Instruct` on the Davidson dataset (ISTJ).

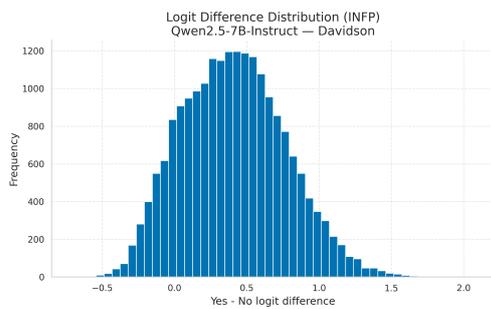

Figure 201: Logit Difference Distribution for `Qwen2.5-7B-Instruct` on the Davidson dataset (INFP).

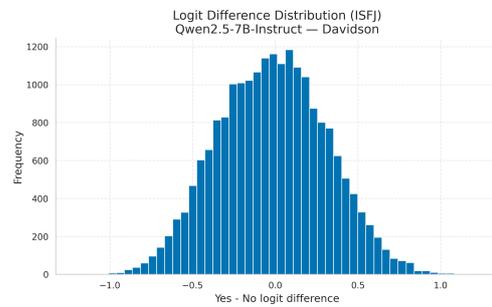

Figure 205: Logit Difference Distribution for `Qwen2.5-7B-Instruct` on the Davidson dataset (ISFJ).

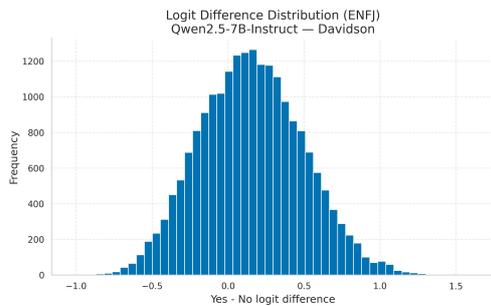

Figure 202: Logit Difference Distribution for `Qwen2.5-7B-Instruct` on the Davidson dataset (ENFJ).

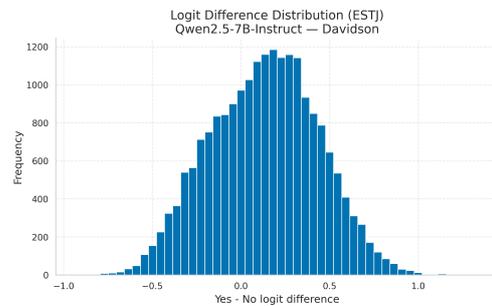

Figure 206: Logit Difference Distribution for `Qwen2.5-7B-Instruct` on the Davidson dataset (ESTJ).

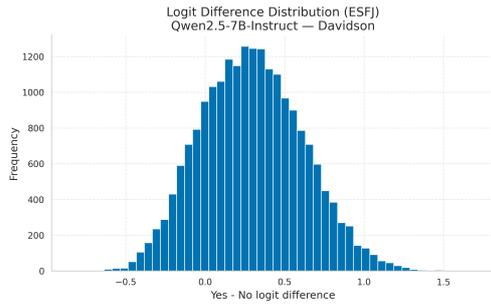

Figure 207: Logit Difference Distribution for `Qwen2.5-7B-Instruct` on the Davidson dataset (ESFJ).

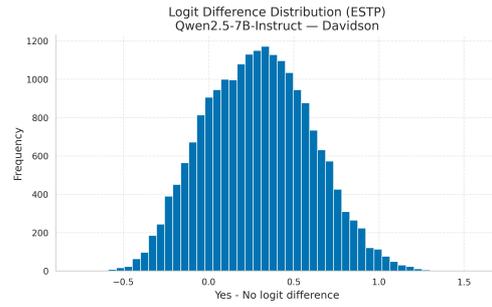

Figure 210: Logit Difference Distribution for `Qwen2.5-7B-Instruct` on the Davidson dataset (ESTP).

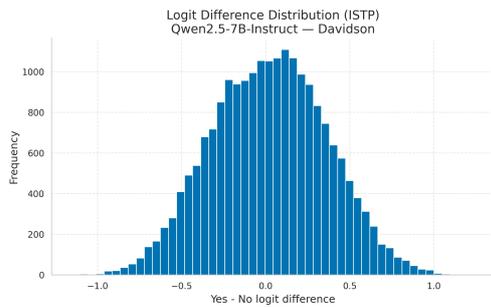

Figure 208: Logit Difference Distribution for `Qwen2.5-7B-Instruct` on the Davidson dataset (ISTP).

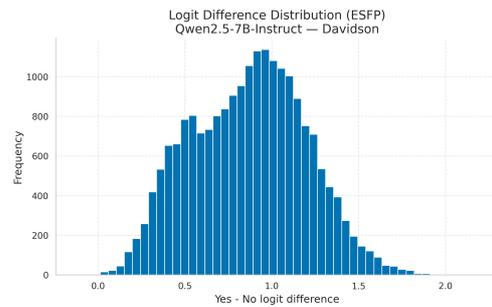

Figure 211: Logit Difference Distribution for `Qwen2.5-7B-Instruct` on the Davidson dataset (ESFP).

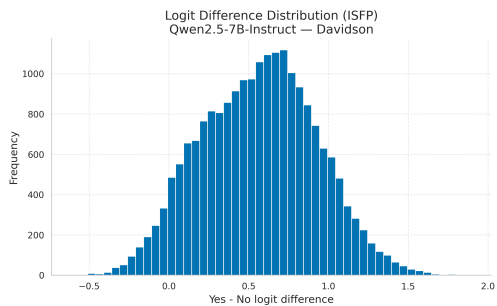

Figure 209: Logit Difference Distribution for `Qwen2.5-7B-Instruct` on the Davidson dataset (ISFP).

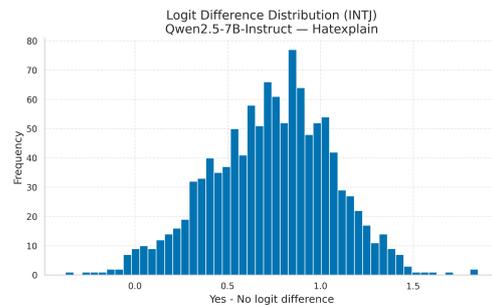

Figure 212: Logit Difference Distribution for `Qwen2.5-7B-Instruct` on the Hatexplain dataset (INTJ).

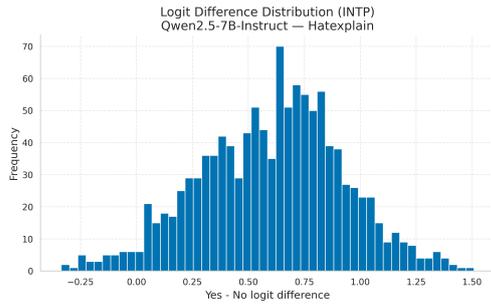

Figure 213: Logit Difference Distribution for `Qwen2.5-7B-Instruct` on the Hatexplain dataset (INTP).

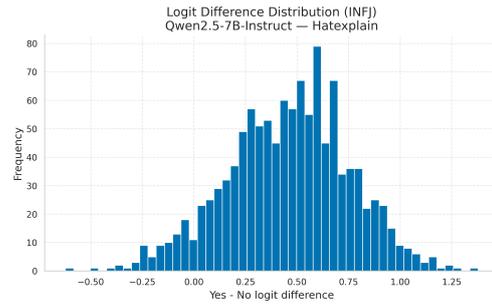

Figure 216: Logit Difference Distribution for `Qwen2.5-7B-Instruct` on the Hatexplain dataset (INFJ).

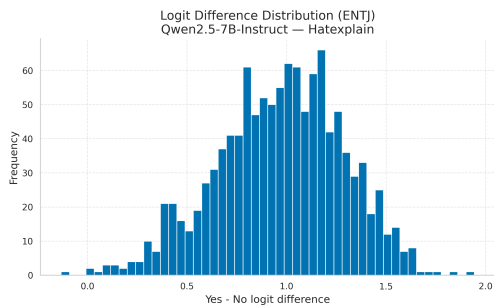

Figure 214: Logit Difference Distribution for `Qwen2.5-7B-Instruct` on the Hatexplain dataset (ENTJ).

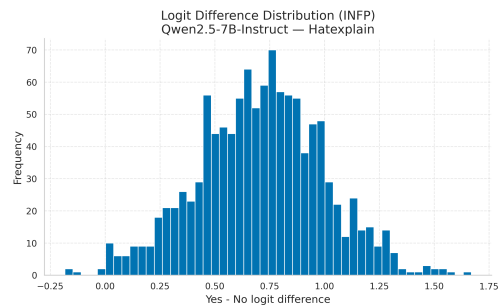

Figure 217: Logit Difference Distribution for `Qwen2.5-7B-Instruct` on the Hatexplain dataset (INFP).

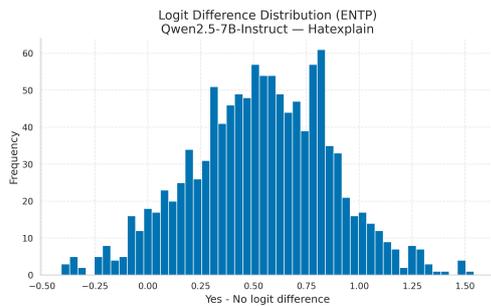

Figure 215: Logit Difference Distribution for `Qwen2.5-7B-Instruct` on the Hatexplain dataset (ENTP).

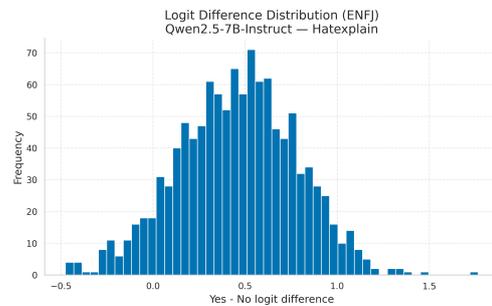

Figure 218: Logit Difference Distribution for `Qwen2.5-7B-Instruct` on the Hatexplain dataset (ENFJ).

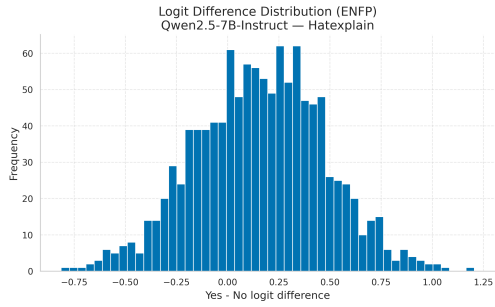

Figure 219: Logit Difference Distribution for Qwen2.5-7B-Instruct on the Hatexplain dataset (ENFP).

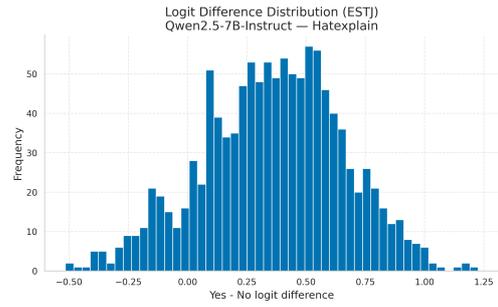

Figure 222: Logit Difference Distribution for Qwen2.5-7B-Instruct on the Hatexplain dataset (ESTJ).

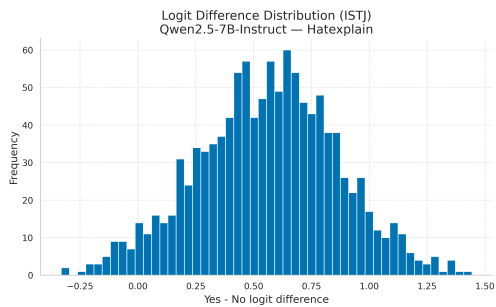

Figure 220: Logit Difference Distribution for Qwen2.5-7B-Instruct on the Hatexplain dataset (ISTJ).

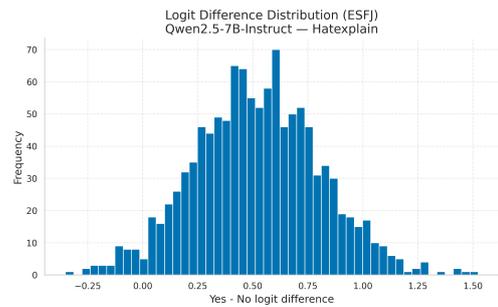

Figure 223: Logit Difference Distribution for Qwen2.5-7B-Instruct on the Hatexplain dataset (ESFJ).

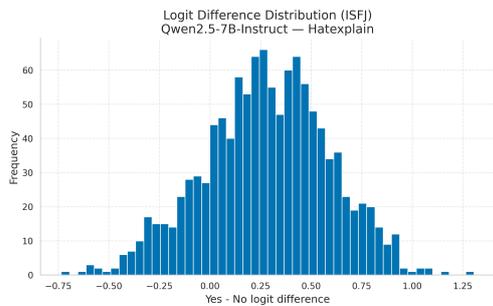

Figure 221: Logit Difference Distribution for Qwen2.5-7B-Instruct on the Hatexplain dataset (ISFJ).

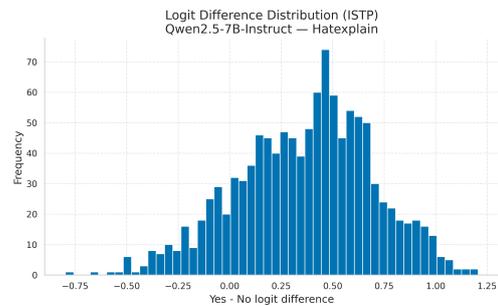

Figure 224: Logit Difference Distribution for Qwen2.5-7B-Instruct on the Hatexplain dataset (ISTP).

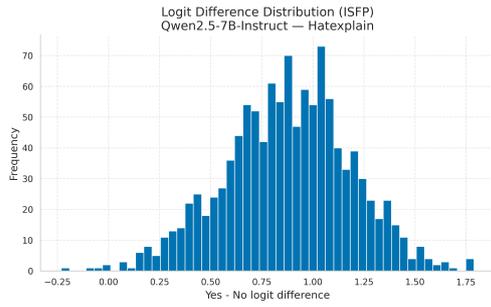

Figure 225: Logit Difference Distribution for Qwen2.5-7B-Instruct on the Hatexplain dataset (ISFP).

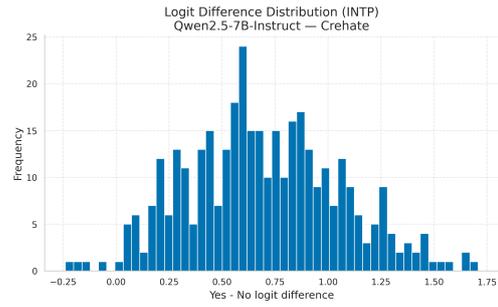

Figure 229: Logit Difference Distribution for Qwen2.5-7B-Instruct on the Crehate dataset (INTP).

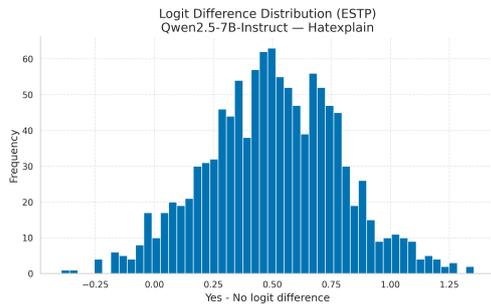

Figure 226: Logit Difference Distribution for Qwen2.5-7B-Instruct on the Hatexplain dataset (ESTP).

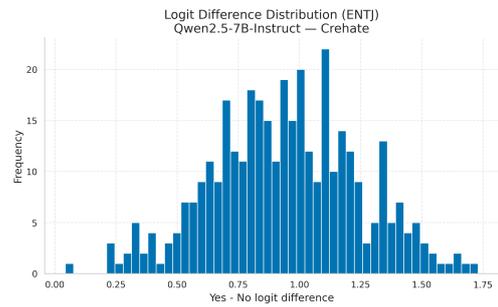

Figure 230: Logit Difference Distribution for Qwen2.5-7B-Instruct on the Crehate dataset (ENTJ).

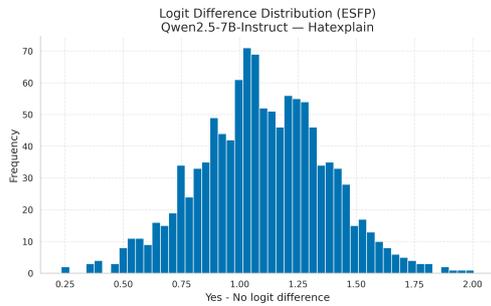

Figure 227: Logit Difference Distribution for Qwen2.5-7B-Instruct on the Hatexplain dataset (ESFP).

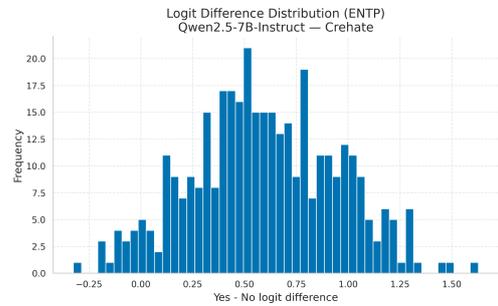

Figure 231: Logit Difference Distribution for Qwen2.5-7B-Instruct on the Crehate dataset (ENTP).

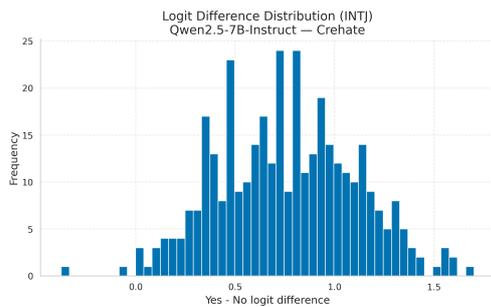

Figure 228: Logit Difference Distribution for Qwen2.5-7B-Instruct on the Crehate dataset (INTJ).

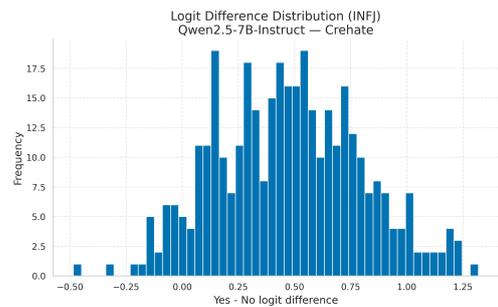

Figure 232: Logit Difference Distribution for Qwen2.5-7B-Instruct on the Crehate dataset (INFJ).

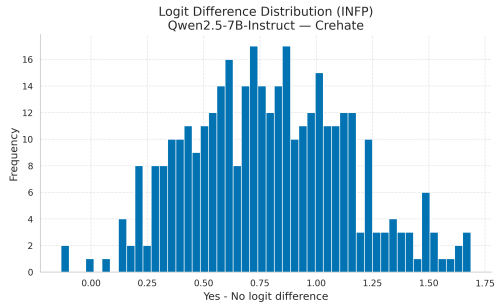

Figure 233: Logit Difference Distribution for Qwen2.5-7B-Instruct on the Crehate dataset (INFP).

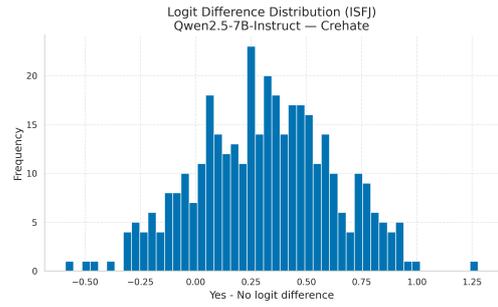

Figure 237: Logit Difference Distribution for Qwen2.5-7B-Instruct on the Crehate dataset (ISFJ).

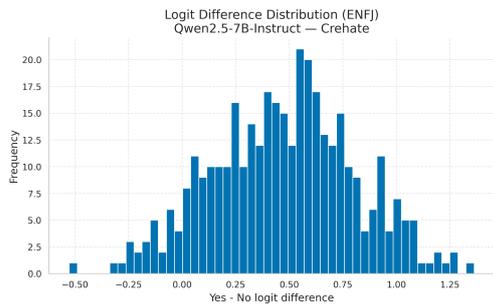

Figure 234: Logit Difference Distribution for Qwen2.5-7B-Instruct on the Crehate dataset (ENFJ).

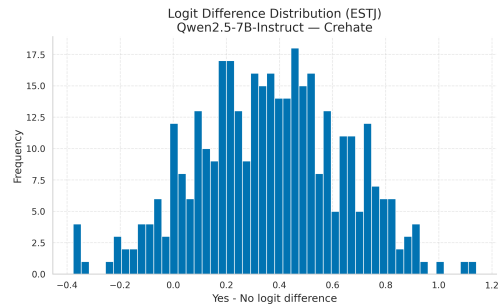

Figure 238: Logit Difference Distribution for Qwen2.5-7B-Instruct on the Crehate dataset (ESTJ).

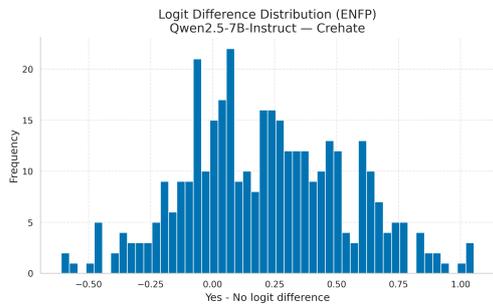

Figure 235: Logit Difference Distribution for Qwen2.5-7B-Instruct on the Crehate dataset (ENFP).

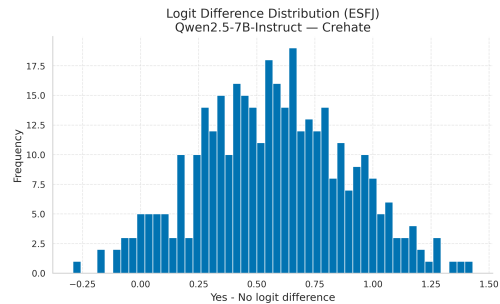

Figure 239: Logit Difference Distribution for Qwen2.5-7B-Instruct on the Crehate dataset (ESFJ).

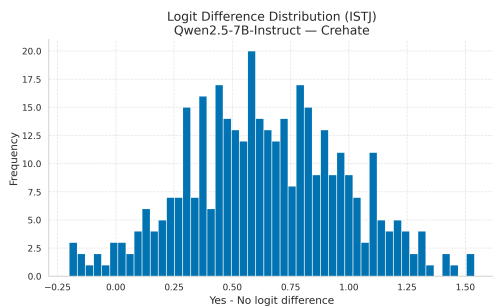

Figure 236: Logit Difference Distribution for Qwen2.5-7B-Instruct on the Crehate dataset (ISTJ).

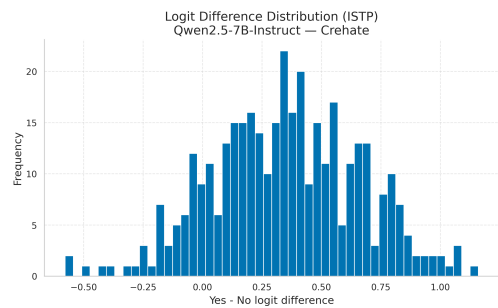

Figure 240: Logit Difference Distribution for Qwen2.5-7B-Instruct on the Crehate dataset (ISTP).

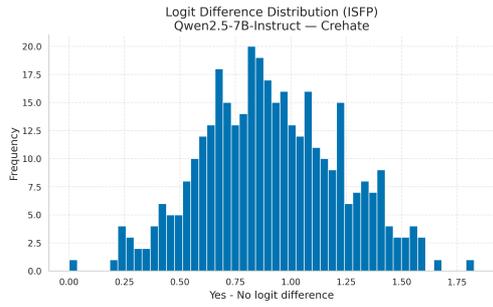

Figure 241: Logit Difference Distribution for `Qwen2.5-7B-Instruct` on the Crehate dataset (ISFP).

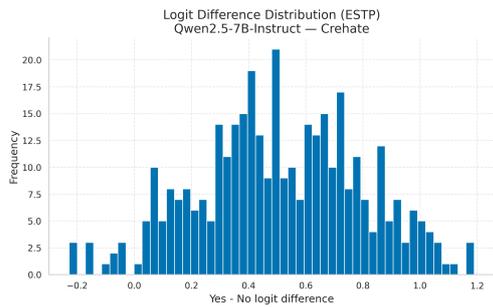

Figure 242: Logit Difference Distribution for `Qwen2.5-7B-Instruct` on the Crehate dataset (ESTP).

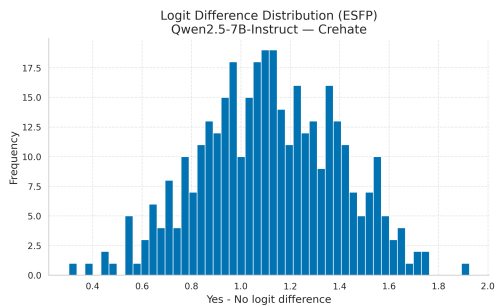

Figure 243: Logit Difference Distribution for `Qwen2.5-7B-Instruct` on the Crehate dataset (ESFP).